%% file: main.tex
\documentclass[10pt,twocolumn,letterpaper]{article}

\usepackage{wacv} 
\input{preamble}

\definecolor{wacvblue}{rgb}{0.21,0.49,0.74}
\usepackage[pagebackref,breaklinks,colorlinks,allcolors=wacvblue]{hyperref}
\def\wacvPaperID{267} % *** Enter the WACV Paper ID here
\def\confName{WACV}
\def\confYear{2026}

\title{\textcolor{orange}{\text{\faBolt}}FLARES\textcolor{orange}{\text{\faBolt}}: Fast and Accurate LiDAR Multi-Range Semantic Segmentation}

\author{
Bin Yang\textsuperscript{1,2} \quad
Alexandru Paul Condurache\textsuperscript{1,2}\\
\textsuperscript{1}Automated Driving Research, Robert Bosch GmbH\\
\textsuperscript{2}Institute for Signal Processing, University of Lübeck\\
{\tt\small \{Bin.Yang3, AlexandruPaul.Condurache\}@de.bosch.com}
}
\begin{document}
\maketitle
\input{sec/0_abstract}    
\input{sec/1_intro}
\input{sec/2_related_works}

\input{sec/3_methods}

\input{sec/4_experiments}
\input{sec/5_conclusion}
\clearpage
{
    \small
    \bibliographystyle{ieeenat_fullname}
    \bibliography{main}
}
\input{sec/X_suppl}
\end{document}

%% file: preamble.tex
\usepackage[dvipsnames]{xcolor}
\usepackage{algorithm}
\usepackage{algpseudocode}
\usepackage[ruled,vlined,algo2e]{algorithm2e}
\usepackage[inkscapelatex=false]{svg}
\usepackage{marvosym}
\usepackage{multirow}
\usepackage{color}
\usepackage{colortbl}
\usepackage[font=small,skip=0pt]{caption}
\usepackage{float}
\usepackage{fontawesome}
\providecommand{\SetAlgoLined}{\SetLine}

\definecolor{LightCyan}{rgb}{0.88, 1.0, 1.0}
\definecolor{SemanticKITTI-Road}{RGB}{255, 0, 255}
\definecolor{SemanticKITTI-Building}{RGB}{255, 200, 0}
\definecolor{SemanticKITTI-Veg}{RGB}{0, 175, 0}
\definecolor{SemanticKITTI-Car}{RGB}{100, 150, 245}
\definecolor{SemanticKITTI-Pole}{RGB}{248, 238, 146}
\definecolor{incorrect}{RGB}{255, 0, 0}
\definecolor{correct}{RGB}{100, 100, 100}
\newcommand{\coolname}{\textit{FLARES}}
\newcommand{\PAR}[1]{\vskip2pt \noindent{\bf #1}}

\newcolumntype{?}{!{\vrule width 1pt}}
\newcolumntype{|}{!{\vrule width .5pt}}

\newcommand{\redcirclednumber}[1]{\raisebox{.5pt}{\textcolor{red}{\textcircled{\raisebox{-.9pt} {#1}}}}}

%% file: sec/0_abstract.tex
\begin{abstract}
3D scene understanding is a critical yet challenging task in autonomous driving due to the irregularity and sparsity of LiDAR data, as well as the computational demands of processing large-scale point clouds. Recent methods leverage range-view representations to enhance efficiency, but they often adopt higher azimuth resolutions to mitigate information loss during spherical projection, where only the closest point is retained for each 2D grid. However, processing wide panoramic range-view images remains inefficient and may introduce additional distortions. Our empirical analysis shows that training with multiple range images, obtained from splitting the full point cloud, improves both segmentation accuracy and computational efficiency. However, this approach also poses new challenges of exacerbated class imbalance and increase in projection artifacts. To address these, we introduce \coolname{}, a novel training paradigm that incorporates two tailored data augmentation techniques and a specialized post-processing method designed for multi-range settings. Extensive experiments demonstrate that \coolname{} is highly generalizable across different architectures, yielding 2.1\%–7.9\% mIoU improvements on SemanticKITTI and 1.8\%–3.9\% mIoU on nuScenes, while delivering over 40\% speed-up in inference.\footnote{Project page: \url{https://binyang97.github.io/FLARES}}
\end{abstract} 

%% file: sec/1_intro.tex
\section{Introduction}
\label{sec:intro}
LiDAR is one of the most common sensors for perception in autonomous driving. Semantic segmentation on LiDAR point clouds is essential for getting useful and reliable information of the surrounding 3D environment. To solve this 3D scene understanding task, many prior works propose to integrate deep learning techniques because of its remarkable advancements in the past few years. The publication of various annotated datasets~\cite{geiger2012we, caesar2020nuscenes, sun2020scalability} in the domain of autonomous driving further promotes research in the field.\\
In general, those methods can be categorized based on LiDAR data representation into point-based~\cite{zhao2021point, qi2017pointnet++, wu2024ptv3}, voxel-based~\cite{zhu2021cylindrical, hong2022dynamic} and projection-based methods~\cite{zhang2020polarnet, 2019rangenet++, cheng2022cenet, zhao2021fidnet}. Both point and voxel-based approaches typically require substantial computational resources due to the need to process data through networks with numerous 3D convolutional layers, intensive feature pre-processing, and deep architectures involving multiple downsampling and upsampling operations. These requirements can result in slow inference speeds, limiting their suitability for real-time applications~\cite{loiseau22online}. In contrast, rasterizing point cloud into range-view images~\cite{cortinhal2020salsanext} is more advantageous in fast and scalable LiDAR perception, because it allows the use of 2D operators for efficient computation and facilitates the transfer of knowledge from camera images~\cite{ando2023rangevit, kong2023rethinking}.
\begin{figure}[t]
% \vspace{-1mm}
\centering
\begin{subfigure}{.48\textwidth}
\centering
    \includesvg[width=\columnwidth]{pics/standard.svg}
    \caption{Standard}
    \label{subfig:standard}
\end{subfigure}
\begin{subfigure}{.48\textwidth}
\centering
    \includesvg[width=\columnwidth]{pics/str.svg}
    \caption{STR~\cite{kong2023rethinking}}
    \label{subfig:str}
\end{subfigure}
\begin{subfigure}{.48\textwidth}
\centering
    \includesvg[width=\columnwidth]{pics/flares.svg}
    \caption{\coolname{}}
    \label{subfig:flares}
\end{subfigure}
\caption{Visual comparison among different training procedures for range-view LiDAR semantic segmentation: \redcirclednumber{1} Splitting, \redcirclednumber{2} Range-view projection, \redcirclednumber{3} Network prediction, \redcirclednumber{4} Post-processing, \redcirclednumber{5} Image concatenation.}
\label{fig:intro}
\vspace{-5mm}
\end{figure}\\
%
% Nevertheless, learning from range-view representations can suffer from the "many-to-one" conflict of adjacent points. The issue cause the irreversible loss in spatial information and thus, leads to sub-optimal performance compared to methods of other 3D LiDAR representations. The loss can be alleviated by enlarging the azimuth resolution of range images, however, it requires the network to process a wide panoramic image (Fig.\ref{subfig:standard}), which significantly increases the computational overhead~\cite{2019rangenet++, cortinhal2020salsanext, cheng2022cenet, zhao2021fidnet, gu2022maskrange, kong2023rethinking}. Building upon this standard setting, recent work advances a scalable training with range-view images (STR) to improve the efficiency~\cite{kong2023rethinking}. Splitting the point clouds into multiple distinct views, STR uses one crops of the range image to train the network and process all crops in a batch during inference. After that, the predictions are jointed to get the full range image. Finally, the full range image is post-processed to 3D coordinates. 
Nevertheless, learning from range-view representations can suffer from a "many-to-one" conflict among adjacent points. This issue causes an irreversible loss of spatial information and leads to inferior performance compared to other methods that utilize 3D LiDAR representations. Although increasing the azimuth resolution of range images can alleviate this loss, it forces the network to process wide panoramic images (Fig.~\ref{subfig:standard}), which significantly increases computational overhead~\cite{2019rangenet++, cortinhal2020salsanext, cheng2022cenet, zhao2021fidnet, gu2022maskrange, kong2023rethinking}. Building on this standard setting, recent work introduces a scalable training framework with range-view images (STR) to improve efficiency~\cite{kong2023rethinking}. As shown in Fig.~\ref{subfig:str}, by splitting the point clouds into multiple distinct views, STR processes all crops in a batch during inference. The resulting predictions are then merged to reconstruct the full range image, which is subsequently post-processed to generate 3D coordinates. \\
However, the splitting strategy of STR is heuristic and compromises scene completeness. Empirical results in~\cite{kong2023rethinking} reveal a slight performance drop within some networks. Furthermore, the gain in inference speed is limited when maintaining the high resolution of range images during the projection and post-processing stages. To solve these challenges, we propose \coolname{}, a novel training paradigm for range-view LiDAR semantic segmentation. Its procedure is visually illustrated in Fig.~\ref{subfig:flares}. In contrast to cartesian grid splits of STR~\cite{kong2023rethinking}, \coolname{} divides point clouds along the LiDAR’s spherical coordinate and then projects each split into lower-resolution range images. This design preserves \textbf{local geometric coherence} and \textbf{scene integrity} during partitioning, which enables \textbf{more effective semantic context sharing} across generated range images compared to STR. Furthermore, our method exclusively processes low-resolution images, achieving \textbf{greater efficiency} than both STR and standard settings. 
%
% \begin{figure}[b]
% \vspace{-8mm}
%   \centering
%     \begin{subfigure}{.49\textwidth}
%     \centering
%         \includesvg[width=\columnwidth]{pics/misalignment.svg}
%         \caption{Elevation-Azimuth plot of a LiDAR point cloud. \textcolor{red}{Misalignment}: points at the same azimuth position are not always vertically aligned.}
%         \label{subfig:misalignment}
%     \end{subfigure}
%   \begin{subfigure}{.4\textwidth}
%     \includesvg[width=\columnwidth]{pics/occupancy.svg}
%     % \includesvg[width=0.9\linewidth]{pics_svg/Occupancy_validity.svg}
%     \caption{Statistics on SemanticKITTI~\cite{behley2019semantickitti}: 3D validity (proportion of projected points) with different azimuth (\textbf{W}) and elevation (\textbf{H}) resolutions. Comparable increases are observable when doubling azimuth and elevation resolution ($\Delta V_{azi}, \Delta V_{ele}$).}
%     \label{subfig:occupancy}
%    \end{subfigure}
%     \vspace{-2mm}
%     \caption{}
% \end{figure}\\
\begin{figure}[b]
\vspace{-6mm}
    \includesvg[width=0.9\columnwidth]{pics/occupancy.svg}
    % \includesvg[width=0.9\linewidth]{pics_svg/Occupancy_validity.svg}
    \caption{Statistics on SemanticKITTI~\cite{behley2019semantickitti}: 3D validity (proportion of projected points) with different azimuth (\textbf{W}) and elevation (\textbf{H}) resolutions. Comparable increases are observable when doubling azimuth and elevation resolution ($\Delta V_{azi}, \Delta V_{ele}$).}
    \label{subfig:occupancy}
    \vspace{-2mm}
\end{figure}\\
Moreover, multi low-resolution range-view images created by \coolname{} are generally a more informative and powerful representation compared to the high-resolution projection. Due to ego-motion and sensor calibration error, points mapped to identical azimuthal positions often misalign with actual laser beam orientations~\cite{yang2022lidar}. This inherent mismatch demonstrates that jointly prioritizing azimuth and elevation resolutions, rather than solely expanding image width, achieves higher projection fidelity. Statistically, as quantified in our occupancy measurement (Fig.~\ref{subfig:occupancy}), doubling the elevation resolution achieves approximately 80\% of the projection gain obtained by doubling the azimuth resolution in SemanticKITTI~\cite{behley2019semantickitti} dataset. This property indicates that our new representation can restore more 3D points, thereby alleviating the "many-to-one" conflict among adjacent points more effectively.
% unlike the naive maximization of azimuth resolution used in standard approaches and STR to alleviate the "many-to-one" problem, \coolname{}’s splitting strategy enhances the informativeness through multi-range projection.
%
\begin{figure}[t]
% \vspace{-4mm}
\centering
\begin{subfigure}{.235\textwidth}
\centering
    \includesvg[width=\columnwidth]{pics/fullcloud.svg}
    \caption{Before splitting}
    \label{subfig:fullcloud}
\end{subfigure}
\begin{subfigure}{.235\textwidth}
\centering
    \includesvg[width=\columnwidth]{pics/subcloud.svg}
    \caption{After splitting}
    \label{subfig:subcloud}
\end{subfigure}
\caption{An example from SemanticKITTI~\cite{behley2019semantickitti} dataset is visualized in top-down view and range-view. Three crops are magnified to specify \textit{three limitations} brought by the splitting of point clouds. \textbf{1) Exacerbated class imbalance}: objects with small sizes, likewise low occurrences in annotations, tend to fade away in the range image after downsampling (e.g. \textcolor{SemanticKITTI-Pole}{pole} in the left crop). \textbf{2) Intensified noise}: reduction in points results in increasing clutters (middle crop) that may disrupt the training stability. \textbf{3) Deteriorated distortion}: decrease in point density can introduce more projection artifacts, thus corrupting the sharpness of local geometry (e.g. blurry boundaries of \textcolor{SemanticKITTI-Car}{car} in the right crop).}
\vspace{-5mm}
\label{fig:drawbacks_splitting}
\end{figure}\\
While \coolname{} offers significant advantages, it introduces three key challenges, as visualized in Fig.~\ref{fig:drawbacks_splitting}. First, point cloud splitting exacerbates class imbalance issues~\cite{classimbalance}, where classes representing small objects with inherently sparse distributions become further undersampled. Second, the splitting operation increases point cloud sparsity, causing substantial reductions in occupancy of range images. These unoccupied pixels act as noise that destabilize training convergence~\cite{kong2023rethinking}. Finally, the resulting projection artifacts degrade image sharpness, particularly at partition boundaries where geometric discontinuities emerge. \\
To tackle these incidental issues, we extend the pipeline with two data augmentation steps. Furthermore, we investigate in exploring a more effective post-processing technique tailored to the multi-range settings to further enhance the advantages of our method.\\
In summary, our contributions are: 
\begin{enumerate}[noitemsep,topsep=0pt,itemsep=0pt]
    \item \coolname{}, a scalable training paradigm that enhances LiDAR semantic segmentation with improved speed, accuracy, and architectural agnosticism.
    \item  Two specialized data augmentation techniques addressing exacerbated class imbalance and noise amplification within \coolname{}.
    \item A novel interpolation-based post-processing method that refines range-view predictions using precise nearest-neighbor information.
    \item Comprehensive experiments across four network architectures and two benchmarks demonstrating consistent superiority over baselines.
\end{enumerate}

%% file: sec/2_related_works.tex
\section{Related Works}
\label{sec:related_works} 
\PAR{Point and Voxel-based methods} Some recent works~\cite{qi2017pointnet++, zhao2021point, wu2022ptv2} use raw point cloud data as direct network input, eliminating the need for post-processing after prediction. However, these methods often face high computational complexity and memory usage. To address these issues, \citet{hu2020randla} introduced sub-sampling and feature aggregation techniques for large-scale point clouds to reduce computational costs. Despite these efforts, performance degradation remains significant. Other works~\cite{zhu2021cylindrical, lai2023spherical} use 3D voxel grids as input, achieving point-based accuracy with reduced computational costs by utilizing sparse 3D convolutions~\cite{choy2019minkowski}. Nonetheless, voxelization and de-voxelization steps continue to be time- and memory-intensive.
\PAR{Range-view-based methods} To address inefficiencies, some prior works~\cite{wu2018squeezeseg, 2019rangenet++, wu2019squeezesegv2} convert the large-scale point cloud to panoramic range image through spherical projection and leverage image segmentation techniques for LiDAR data. SalsaNext~\cite{cortinhal2020salsanext} uses a Unet-like network with dilated convolutions to broaden receptive fields for more accurate segmentation, while Lite-HDSeg~\cite{razani2021lite} introduces an efficient framework using a lite version of harmonic convolutions. Additionally, FIDNet~\cite{zhao2021fidnet} and CENet ~\cite{cheng2022cenet} interpolate and concatenate multi-scale features with a minimal decoder for semantic prediction. These methods share the benefit of lightweight network design, significantly improving efficiency and enabling real-time applications. Nevertheless, they generally underperform 3D methods due to the “many-to-one” issue, where multiple points project to the same pixel. To offset the performance drop caused by the problem, some other recent works propose to use Vision Transformer (ViT)~\cite{dosovitskiy2020image, xie2021segformer, wang2021pyramid}. RangeViT~\cite{ando2023rangevit} deploys standard ViT backbone as encoder, followed by a light-weight decoder for refining the coarse patch-wise ViT representations, while RangeFormer~\cite{kong2023rethinking} utilizes a pyramid-wise ViT-encoder to extract multi-scale features from range images. ViTs offer higher model capacities and excel at capturing long-range dependencies by modeling global interactions between different regions, enhancing segmentation performance over traditional CNNs~\cite{deininger2022comparative}. However, the quadratic computational complexity of self-attention mechanisms in ViTs introduces challenges in achieving an optimal balance between efficiency and accuracy.
% In contrast to previous approaches, we reassess the necessity of high-resolution range images and develop a customized framework to efficiently and accurately process range-view data while maintaining computational feasibility.
% recent works~\cite{cheng2022cenet, cortinhal2020salsanext} use higher-resolution range images. While this improves performance, it also increases model complexity and reduces inference speed, counteracting the advantages of range-view methods. Driven by the success of ViTs~\cite{dosovitskiy2020image} in the field of RGB image segmentation~\cite{xie2021segformer, wang2021pyramid}, ViTs have been extended to range-view semantic segmentation of LiDAR point clouds.
%
\PAR{Training Paradigm} While existing methods address inefficiencies in high-resolution range images by employing compact networks to reduce model capacity~\cite{zhao2021fidnet, cortinhal2020salsanext, cheng2022cenet}, these approaches still require substantial memory for processing high-resolution inputs, limiting scalability in batch size and data throughput. Reducing resolution mitigates memory demands but worsens spatial information loss, degrading segmentation accuracy. To resolve this trade-off, \citet{kong2023rethinking} introduced Scalable Training from Range-view (STR), which splits range images into sub-images from multiple perspectives to lower memory consumption. However, STR’s use of partial scene views during training inherently limits segmentation accuracy. Furthermore, the method does not optimize for computational efficiency, leaving runtime improvements unaddressed.
 % \citet{cheng2022cenet} demonstrated that increasing the azimuth resolution by a factor of four can reduce inference speed by more than half. 
% \PAR{Augmentation} Data augmentation plays a crucial role in helping models learn more generalized representations, thereby enhancing scalability. For example, Mix3D~\cite{nekrasov2021mix3d} introduced an out-of-context mixing strategy by fusing two scenes, while PolarMix~\cite{xiao2022polarmix} generates a new training sample through blending polar sections of two point clouds. Similarly, MaskRange~\cite{gu2022maskrange} proposed a weighted paste-drop augmentation to manually fuse instance-level semantic contexts from other scenes. RangeFormer~\cite{kong2023rethinking} employed four consecutive range-wise operations to provide richer semantic and structural cues in the scene. Unlike these generalization-focused methods, we innovate two framework-specific augmentations tailored to address the unique challenges of \coolname{}.
%Similarly, MaskRange~\cite{gu2022maskrange} proposed a weighted paste-drop augmentation to manually balance the class frequencies,
\PAR{Post-Processing} Addressing the prevalent "many-to-one" problem in range-view representations often necessitates a post-processing step to upsample 2D predictions, a critical yet under-explored area in prior research. RangeViT~\cite{ando2023rangevit} introduced a trainable 3D refiner using KPConv~\cite{thomas2019kpconv}. However, while it directly optimizes 3D semantics, the performance improvement is limited, and the approach adds significant computational overhead. Some methods rely on conventional unsupervised techniques to infer semantics for 3D points. For instance, \citet{2019rangenet++} proposed a KNN-based voting approach, and \citet{zhao2021fidnet} introduced Nearest Label Assignment (NLA), which assigns labels based on the closest labeled point in 3D space. Nevertheless, these unsupervised techniques often struggle with accurately predicting boundaries and distant points, with performance further declining as range-image resolution decreases. To overcome these limitations, we design a new post-processing method that adaptively weights neighboring contributions for interpolating predictions in 3D space.

% We hypothesize that this limitation stems from the lack of depth information, which restricts the method's ability to accurately reconstruct the 3D feature space from 2D features.
% \PAR{Range-view-based methods (ViTs)} In addition, RangeViT~\cite{ando2023rangevit} propose to replace the conventional KNN-based post-processing approaches with KPconv~\cite{thomas2019kpconv} layers to make the back-projection process trainable.
% \PAR{Hybrid methods} Some other works combine the strengths of point-, projection-, and voxel-based approaches. SPVNAS~\cite{tang2020searching} uses a sparse point-voxel convolution with a point-based branch for local features and a voxel-based branch for global context. RPVNet~\cite{xu2021rpvnet} further integrates these approaches with three branches—point, projection, and voxel—using the point-based branch to fuse information from the others. The 

%% file: sec/3_methods.tex
\section{Methodology}
\subsection{Proposed Framework}
% In response to those limitations, we propose a novel framework for range-view semantic segmentation that is generalizable across different networks and significantly improves performance and efficiency compared to their baselines. 
% Our method introduces key optimizations in separate training and inference scheme for low-resolution range images, an enhanced data augmentation techniques, and a novel effective post-precossing approach for mapping 2D prediction into 3D space. %The following sections provide a detailed technical explanation of our approach.

% Our method introduces critical optimizations across three areas: a distinct training and inference scheme for low-resolution range images, advanced augmentation techniques, and an effective post-processing approach for accurately mapping 2D predictions into 3D space.
%
A LiDAR point cloud consists of points captured during a single revolution, denoted as \( P = \{p_1, ..., p_n\} \), where each measurement represents a 4D point including the cartesian coordinates \( p_i = \{x_i, y_i, z_i\} \) and intensity \(t_i\). In semantic segmentation task, each point has an additional feature of class label, denoted as $c_i$, for training. Our splitting configuration operates as follows: given a predefined partition count $N$, we distribute points into sub-clouds by along the spherical coordinate of the scan pattern. Specifically, a sub-cloud is derived as $P_{i} = \{p_j| j \mod N = i-1\}$ for $i\in\{1,2, ..N\}$. This modulo-based assignment ensures equal partitioning while preserving spatial coherence within each sub-cloud. For the spherical projection, we specify the function $f$ as following: 
\begin{equation} \label{eqn:prediction}
f(p) =
\begin{bmatrix} u \\ v \end{bmatrix} = 
\begin{bmatrix} \frac{W}{2}  - \frac{W}{2\pi} arctan(\frac{y}{x}) \\
 \frac{H}{\Theta_{max}-\Theta_{min}} * (\Theta_{max} - arcsin(\frac{z}{d}))
\end{bmatrix}
\end{equation}
, where \(W\) and \(H\) as the image width and height, while \(v\) and \(u\) correspond to the elevation and azimuth angles of LiDAR points. Angular values $\Theta_{max}$ and $\Theta_{min}$ define the upper and lower bound of the LiDAR's vertical field of views and the depth value is calculated by $d = \sqrt{x^2 + y^2 + z^2}$. Note that \(H\) is typically determined by the number of LiDAR sensor beams, while \(W\) can be assigned with random value based on the requirements. Similar to prior studies~\cite{cheng2022cenet, zhao2021fidnet, cortinhal2020salsanext}, we adopt a five-channel input representation (\(x, y, z, t, d\)). \\
Next, we project all sub-clouds into range images simultaneously $R_i = f(P_i)$. During \textit{training}, since our splitting strategy leverages shared semantic contexts across sub-clouds, we randomly select \textit{only one} for network optimization. During \textbf{inference}, all images $R_i$ are \textbf{stacked as a batch} and processed by the network to extract the 2D predictions. Unlike STR~\cite{kong2023rethinking}, which wraps all images to the original resolution, we treat each sub-cloud as an independent complete scene and apply post-processing to low-resolution range images individually. Final 3D predictions are reconstructed by merging outputs from all sub-clouds.\\
An discussed in Sec.~\ref{sec:intro}, \coolname{} offers three key advantages: 1) Enhanced restoration of original 3D information through multi-range projection, 2) Reduced memory consumption by decreasing the image resolution, enabling the scalable training on smaller GPUs, 3) Preservation of the full field of view, which maintains contextual integrity despite downsampling.

% To balance 3D validity and 2D occupancy, assuming $N_{max}$ is the maximum number of partitioning groups for the specific resolution of range images, it is determined by the rule that the average 2D occupancy must not fall below the high-resolution range image in standard mode ($\frac{1}{N_{max}}\sum^{N_{max}}_i Occ_{i} \geq Occ_{high}$). This new design offers three key advantages: \textbf{1)} Enhanced projection rate by increasing both image height and width; \textbf{2)} Reduced memory consumption, enabling deployment on smaller GPUs; \textbf{3)} Preservation of the full field of view, maintaining contextual integrity despite downsampling.

\subsection{Data Augmentation}
\label{subsec:da}
In particular, the point cloud splitting strategy of FLARES implicitly introduces two new challenges: exacerbated class imbalance and amplified projection noise. To address class imbalance, prior works~\cite{gu2022maskrange, xiao2022polarmix, xu2021rpvnet} have primarily proposed augmenting long-tail classes by pasting them from other frames. While effective to some extent, we find that the augmentation strength is insufficient to counteract the heightened imbalance in our setting. To this end, we propose an enhanced data augmentation strategy specifically designed for class balancing. As for projection noise, recent approaches~\cite{xu2023frnet} attempt to mitigate it by interpolating pseudo pixels into unoccupied regions. However, this often introduces new artifacts in the range-view image. In contrast, tailored to the multi-range configuration of \coolname{}, we leverage the unused sub-clouds to increase pixel occupancy in the sampled range-view image, thereby reducing projection artifacts. 
% Previous studies~\cite{cheng2022cenet, ando2023rangevit, zhao2021fidnet} have primarily used geometric transformations to improve the model generalization such as random flipping, translation, and rotation for point cloud augmentation.  However, this is not sufficient within our framework. In particular, point cloud splitting of \coolname{} implicitly poses two new challenges: exacerbated class imbalance and amplified projection noise. These issues can negatively affect performance, potentially offsetting the benefits of our approach. To address these challenges, we propose two additional data augmentation steps.
\PAR{Weighted Paste-Drop+ (WPD+)} To address the problem of class imbalance, a pervasive challenge in LiDAR semantic segmentation benchmarks~\cite{behley2019semantickitti, caesar2020nuscenes}, Weighted Paste-Drop (WPD)~\cite{gu2022maskrange} technique provides a simple yet effective solution by selectively pasting pixels from rare classes while dropping pixels from abundant ones. Building on this approach, we introduce WPD+, a fundamentally re-engineered augmentation strategy specifically designed for \coolname{}.\\
Unlike traditional methods that apply augmentation uniformly on range-view images~\cite{gu2022maskrange, kong2023rethinking}, WPD+ operates directly in 3D space, This design not only avoids repeated geometric transformations on sampled and current frames, but also harnesses multiple scenes to more effectively balance class distributions and mitigate the severe imbalance induced by point cloud splitting. \\
To further enhance occurrence of underrepresented classes, particularly those corresponding to small and dynamic objects in the scene, we incorporate a curated synthetic dataset from the Carla Simulator~\cite{dosovitskiy2017carla} (more details are provided in Appendix A.5). From our empirical results, despite inherent domain gaps, this approach has led to notable accuracy gains in long-tail classes.
% By integrating these innovations into a unified training framework, WPD+ delivers significant performance by 

% Class imbalance is a common issue in LiDAR semantic segmentation benchmarks~\cite{behley2019semantickitti, caesar2020nuscenes}, where certain classes are heavily underrepresented. This imbalance is compounded by information loss during point cloud-to-range image conversion, especially when downscaling azimuth resolution. Building upon the Weighted Paste Drop (WPD) from the prior work~\cite{gu2022maskrange}, which pastes pixels from rare classes and drops pixels from abundant classes, we present an enhanced version, WPD+, to mitigate the exacerbation of class imbalance within our framework. Unlike the original method, which performs geometric data augmentation identically on both sampled and current frames in 3D space before projection, our approach applies WPD directly in 3D space, which avoids the repeating computation of geometric transformations, and samples multiple frames to improve class balancing. Additionally, we use a small set of synthetic dataset generated in the Carla Simulator~\cite{dosovitskiy2017carla} to further augment rare classes that correspond to small and dynamic objects in the scene. Despite possible domain gaps between datasets, it yields notable accuracy improvements from our experimental results. 
% %
\PAR{Multi-Cloud Fusion (MCF)}  Due to the point cloud splitting in \coolname{}, the decrease in 2D occupancy leads to intensified projection artifacts in range images. To address the issue, we propose MCF, a strategy tailored to multi-range settings of \coolname{}. We formalize this process as follows. Let \( R_i: \Omega \to \mathbb{R} \cup \{\varnothing\} \) denote the range image from the \( i \)-th sub-cloud over the pixel domain \( \Omega \). For each pixel \( x \in \Omega \), We first define the set using other range images:
% Given the inherent sparsity of LiDAR point clouds, range images often contain a considerable number of empty grids.
\[
S(x) = \{\, R_j(x) \mid j \neq i \text{ and } R_j(x) \neq \varnothing \,\}.
\]
Then, for a range image \( R_i \) over the pixel domain \( \Omega \), the occupancy-filled value \( \tilde{R}_i(x) \) is given by:
\[
\tilde{R}_i(x) =
\begin{cases}
R_i(x), & \text{if } R_i(x) \neq \varnothing, \\[1ex]
f(S(x)), & \text{if } R_i(x) = \varnothing \text{ and } S(x) \neq \emptyset, \\[1ex]
\varnothing, & \text{otherwise.}
\end{cases}
\]
Here, \( f(\cdot) \) is an aggregation function that combines the occupied pixel values from the remaining \( N-1 \) range images. Consequently, this method maximizes the 2D occupancy in the range image of a sub-cloud while maintaining the structural consistency of the scene.\\
Despite their simplicity, both augmentation strategies are effective in enhancing segmentation performance. Moreover, when combined with the multi-range representations of \coolname{}, they significantly outperform prior methods (see Appendix B.3 for related experiments).

% To investigate the potential limitations, we conducted empirical studies. In Fig.~\ref{exp}, training on the full point cloud but using sub-clouds for inference resulted in a performance drop in 2D evaluations. Conversely, directly training on sub-clouds yielded sub-optimal results. We hypothesize that this is due to the reduced occupancy when splitting the full point cloud into sub-clouds.

% Besides physically adjusting the class frequencies for more robust training, we explore another data augmentation alongside the sub-clouds training scheme introduced in ~\ref{subsec:preprocessing}. 

\subsection{Post-Processing}
\label{sec:post-processing}
After images are processed by the network, their 2D predictions must be reprojected into 3D space using a post-processing technique. Conventional methods~\cite{2019rangenet++, zhao2021fidnet} typically rely on hard voting from nearest neighbors to infer predictions for points. However, they are designed for the single range image and requires an iterative processing of all images when working on \coolname{}, which results in increased computational complexity. Furthermore, these approaches are limited in their ability to appropriately weight the contributions of each neighbor in 3D coordinates, thus leading to sub-optimal performance. To leverage the advantages of multi-range settings and overcome these limitations, we propose a novel algorithm called \textit{Nearest Neighbors Range Interpolation (NNRI)}. Its pseudo-code is shown in Algo.~\ref{algo:NNRA}. 
% To align with the new inference framework with stacked predictions, we first propose an extension of the standard KNN method~\cite{2019rangenet++}, termed ~\textit{KNN Ensembling}. In the post-processing phase, all sub-clouds are iteratively processed with KNN and votes are ensembled for every point from the full cloud to obtain final predictions.
%
\begin{algorithm}[h]
    % \vspace{-2mm}
	\SetAlgoLined
    \SetKwInOut{Define}{Define}\SetKwInOut{Input}{Input}\SetKwInOut{Output}{Output}
    \Define {$N$ sub-clouds.\\
            The annotation contains $C$ classes}
    \Input{Range images $R_{ranges}$ with size $N \times H \times W$,\\
           Softmax scores $I_{scores}$ with size $N\times C \times H \times W$,\\
           Arrays $R_{all}(p)$ with range values for all points,\\
           Image coordinates $(u_{all}, v_{all})$ for all points, \\
           Kernel size $k$, \\
           Padding $pad$, \\
           Cut-off factor $\alpha$, \\
           Mean of all range values $r_{mean}$,\\
           Standard Deviation of all range values $r_{std}$
           }
	\Output{Array $Labels$ with predicted labels for all points.}

    % 
    % \gets \forall (h_n,w_m)$ where $(h_n, w_m)$ is in the $k \times k$ local patch centered at $(h,w)$ \\

        % \end{algorithmic}
	\BlankLine
    \begin{algorithmic}[1]
    
    \State \textbf{Unfold scores and ranges with $k \times k$ kernel:}
        \Statex \hspace{1em} $S_s(n,h,w,k) \gets \texttt{unfold}(I_{scores}, k, pad)$
        \Statex \hspace{1em} $S_r(n,h,w,k) \gets \texttt{unfold}(R_{ranges}, k, pad)$
    \State \textbf{Extract nearest-neighbors for each point $p$:}
        \Statex \hspace{1em} $\mathbf{N}_s(n,p, k) \gets S_s(n,h,w,k)[..., u_{all}, v_{all}]$
        \Statex \hspace{1em} $\mathbf{N}_r(n,p, k) \gets S_r(n,h,w,k)[..., u_{all}, v_{all}]$
    
    \State \textbf{Compute relative depths: }
        \Statex \hspace{1em} $\mathbf{N}_{rel}(n,p,k) \gets ||(\mathbf{N}_r(n,p, k) - R_{all}(p)||$
    \State \textbf{Compute the cut-off value for each point $p$: }
        \Statex \hspace{1em}  $D(p) = \texttt{exp}(\frac{R(p) - r_{mean}}{r_{std}}) * \alpha$
     \State \textbf{Filter the valid neighbors and compute weights: }
        \Statex \hspace{-1em} $\mathbf{N}_{valid}(n,p,k) \gets \texttt{clamp}(\mathbf{N}_{rel}(n,p, k), \texttt{max} = D(p)$)
        \Statex \hspace{1em} $W(n, p, k) = 1 - \texttt{Normalize}(\mathbf{N}_{valid}(n, p, k))$

    \State \textbf{Weighted Sum for 3D Projection:}
        \Statex \hspace{1em} $Scores(p) = \sum_{i}^{k^2\times n} W(n,p, k) * \mathbf{N}_s(n,p, k)$
    \Statex \hspace{1em} $Labels = \texttt{argmax}_{c\in C}(Scores(p))$
    \State \textbf{Return} $Labels$
    
    \end{algorithmic}
    	\caption{Nearest Neighbors Range Interpolation}\label{algo:NNRA}
\end{algorithm} \\
% \vspace{-1mm}
After applying softmax to the network output, we begin by kernelizing 2D predictions and range images using a pre-defined kernel size ($3\times3$ in our experiments). Next, we assign each point's nearest neighbors in 2D space with corresponding 2D coordinates and stack them along the sub-cloud dimension. The relative depth between each point and its neighbors is computed by taking the absolute difference in depth values. To extract valid data for interpolation, a threshold is needed to filter out distant neighbors. According to the prior knowledge~\cite{hu2022pointdensity, lawin2018densityregistation}, using a constant threshold is sub-optimal due to differing point densities in LiDAR data: closer points are more likely to be affected by outliers due to high density, while farther points struggle to find valid neighbors due to sparsity. To fit this underlying geometry, the range value of each point is employed to determine its cut-off value. By normalizing the range using pre-computed mean and standard deviation, the cut-off value is derived from an exponential function, which approximates the relationship between point-sensor distance and density~\cite{liu2024extend}. This approach simplifies computation by avoiding the costly nearest neighbor search in 3D space to calculate exact density values and adaptively assigns a threshold to each point. Once valid nearest neighbors are identified, they are normalized within the range of $[0, 1]$ to compute interpolation weights. Finally, softmax scores of all 3D points are interpolated by the weighted sum of their nearest neighbors. \\
%
% , where point density decreases exponentially with distance from the sensor~
Overall, NNRI is designed to effectively mitigate the "many-to-one" issue inherent in range-view methods by leveraging distance-wise local neighborhood information in both 2D and 3D. Moreover, the new approach eliminates the need for sub-cloud concatenation, as used in STR~\cite{kong2023rethinking}, and processes all range images in parallel, which significantly reduces the inference time.

%% file: sec/4_experiments.tex
\section{Experimental Analysis}
\subsection{Settings}
\PAR{Datasets }We conduct experiments on two public LiDAR semantic segmentation datasets. \textbf{SemanticKITTI}~\cite{behley2019semantickitti} dataset~\cite{behley2019semantickitti} consists of 22 sequences captured with a 64-beam LiDAR sensor, encompassing 19 semantic classes. The dataset is split as follows: sequences 00 to 10 (excluding 08) are used for training, sequence 08 is reserved for validation, and sequences 11 to 21 are designated for testing. \textbf{nuScenes} dataset~\cite{caesar2020nuscenes} comprises 1,000 driving scenes recorded in Boston and Singapore using a 32-beam LiDAR sensor, leading to a relatively sparse point cloud. After merging similar and infrequent classes, the dataset includes 16 distinct semantic classes.
\PAR{Networks} We revisited prior works and selected three light-weight CNN-based networks: FIDNet~\cite{zhao2021fidnet}, SalsaNext~\cite{cortinhal2020salsanext} and CENet~\cite{cheng2022cenet}) for integration. To further test the effectiveness across heterogeneous architectures, we additionally deploy RangeViT~\cite{ando2023rangevit}, a network composed of a series of Vision Transformer blocks~\cite{dosovitskiy2020image}, in the experimental phase. Original RangeViT uses a trainable KPConv-based 3D projector to get the point-wise predictions. We replace it with our post-processing component to achieve the full integration of our framework and train the model from scratch.
\PAR{Implementation Details }Prior works experimented mostly with the resolution of $64\times2048$ for \textbf{SemanticKITTI}~\cite{cheng2022cenet, 2019rangenet++, cortinhal2020salsanext}, and $32\times960$~\cite{kong2023rethinking} or $32 \times 2048$~\cite{ando2023rangevit} for \textbf{nuScenes}. In contrast, we reduce the azimuth resolution while increasing the projection rate in \coolname{} mode: resolutions of $64 \times 512$ for \textbf{SemanticKITTI} and $32 \times 480$ for \textbf{nuScenes} are fixed for the input and the full point cloud is split into up 3 and 2 sub-clouds during training and inference, respectively. \coolname{} uses the same loss configurations of prior works~\cite{zhao2021fidnet, cheng2022cenet, cortinhal2020salsanext, ando2023rangevit}, including a cross-entorpy loss (focal loss for RangeViT) and Lovász–Softmax loss~\cite{berman2018lovasz}. For training the selected models (excluding RangeViT) on the \textbf{nuScenes} dataset, we standardize the hyperparameter set since no default configurations are provided. Specifically, we use the AdamW optimizer~\cite{loshchilov2017adamw} along with a OneCycle scheduler~\cite{smith2017onecycle}, setting the maximum learning rate to $1e^{-3}$ and training for 150 epochs. All models are trained on four NVIDIA GeForce GTX 1080Ti in distributed mode. 
%

% For our network with Transformer as backbones, we used the AdamW~\cite{loshchilov2017adamw} optimizer and OneCycle~\cite{smith2017onecycle} scheduler with maximum $lr=1e-3$. The models were trained for 60 epochs on the SemanticKITTI dataset and 120 epochs on the nuScenes dataset, respectively. In accordance with ~\cite{ando2023rangevit, gu2022maskrange}, we integrate the Focal loss~\cite{lin2017focal}, Lovasz-Softmax loss~\cite{berman2018lovasz} and boundary loss~\cite{bokhovkin2019boundary} into the main loss function. For reproduction of baseline results and integration of new components,
\subsection{Comparative Study}

\input{tables/qualitative_results.tex}
% Firstly, we compare the results in \coolname{} with the baseline on two datasets. As presented in Tab.~\ref{table:semantickitti_nuscenes}, the performance of all four networks are remarkably improved by trained with new components: 5.3\%, 7.9\% and 3.3\% increase in mIoU on SemanticKITTI~\cite{behley2019semantickitti} and 2.3\%, 3.9\% and 3.1\% on nuScenes~\cite{caesar2020nuscenes} for SalsaNext~\cite{cortinhal2020salsanext}, FIDNet~\cite{zhao2021fidnet} and CENet~\cite{cheng2022cenet}, respectively. Besides, improvments on RangeViT~\cite{ando2023rangevit} with integration of \coolname{} indicates that our method can be generalised to different structure of backbones and boost the segmentation performance significantly.
We compare \coolname{} with baseline models across two datasets. As shown in Tab.~\ref{table:semantickitti_nuscenes}, all four networks see significant improvements: SalsaNext has mIoU gains of 5.3\% on SemanticKITTI and 2.3\% on nuScenes, FIDNet improves by 7.9\% and 3.9\%, and CENet by 3.3\% and 3.1\%. RangeViT, as a ViT-based network, also exhibits huge enhancement in performance, confirming \coolname{}'s generalization across different architectures. This improvement is especially prominent for smaller, dynamic, and under-represented classes such as \textit{truck}, \textit{motorcycle}, \textit{bicycle}, \textit{pedestrian} and \textit{bicyclist}. Notably, with the support of \coolname{}, the network demonstrates improved accuracy in segmenting foreground objects. For instance, as shown in Fig.~\ref{fig:qualitative_results}, our method's predictions exhibit significantly higher correctness compared to the baseline. \\
An exception arises with the \textit{motorcyclist} class in SemanticKITTI, where IoU scores decrease compared to the baseline. Diving into the problem, this can be traced back to the extremely low occurrence of annotations for that class in the dataset. In standard training on low-resolution range images, this class already suffers from poor representation. In \coolname{} mode, the occurrence is further reduced by splitting of the point cloud. This accumulation of downsampling prevents the network from optimizing on that rare class effectively and lead to inferior performance. In contrast, the improvement on nuScenes is more consistent as class frequencies are better balanced. We regard this as a corner case when testing on an class-imbalanced dataset. As a future work to resolve the issue, we aim to explore 3D reconstruction techniques to generate real-world-like pseudo LiDAR point clouds for enhanced augmentation~\cite{chang2024just, manivasagam2020lidarsim}. \\
% \noindent
In Tab.~\ref{tab:SOTA}, we compare the performance of networks boosted by \coolname{} with other state-of-the-art approaches across various modalities. Despite using relatively fewer parameters, \coolname{}-enhanced models achieve segmentation accuracy comparable to other point- or voxel-based methods that deploy much larger and deeper networks. Moreover, our approach significantly outperforms these alternatives in terms of latency, achieving an excellent trade-off between accuracy and efficiency.
\input{tables/semantickitti_test_v2.tex}
\input{tables/val_test.tex}
\subsection{Ablation Study}
To perform the ablation study, we test with CENet~\cite{cheng2022cenet} on \textit{val} set of SemanticKITTI~\cite{behley2019semantickitti} dataset. % For training and inference in \coolname{} mode, the point cloud is split into 3 sub-clouds and projected to multiple range images.
% Finally, we integrate all components into other state-of-the-art networks and compare the results against their respective baselines to assess the generalization capability of the proposed approach.
\input{tables/full_ablation}
\PAR{Component Design} In Tab.~\ref{tab:ablation_full}, we evaluate the contributions of each component in \coolname{}. Starting with the baseline results, we observe that applying STR~\cite{kong2023rethinking} slightly reduces mIoU with the limited improvement in latency. Incorporating \coolname{} into the framework then yields a significant boost in performance compared to the baseline. However, since KNN post-processing must be iteratively applied to multiple range images, the inference speed becomes lower. Next, our two data augmentation methods effectively addressing the exacerbated class imbalance and amplified noise introduced by \coolname{}. Consequently, this leads to a further increase of 1.7\% in mIoU. Finally, replacing the standard post-processing with NNRI not only improves the overall performance but also achieves a remarkable speed-up in inference due to its ability to run parallel computations across multiple range images. Essentially, compared to the baseline performance, we achieve 4.4\% mIoU gain and around 45\% speed-up through the new framework.
\begin{figure}[h]
\centering
\begin{subfigure}{.235\textwidth}
\centering
    \includesvg[width=\columnwidth]{pics/sampled_frames_wpd.svg}
    \caption{}
    \label{subfig:wpd+_frames}
\end{subfigure}
\begin{subfigure}{.235\textwidth}
\centering
    \includesvg[width=\columnwidth]{pics/synthetic_wpd.svg}
    \caption{}
    \label{subfig:synthetic_wpd+}
\end{subfigure}
\caption{a) Ablation on the number of sampled frames for WPD+. b) Comparative plot of IoU scores for top rare classes with and without the synthetic dataset, alongside the class frequency distribution in the validation set. For all inferences, KNN~\cite{2019rangenet++} is used as the post-processing approach.
}
\vspace{-2mm}
% Our WPD+ includes sampling 6 frames from the same dataset and 2 from the synthetic one.
\label{fig:ablation_da}
\end{figure}
\PAR{WPD+} To address the challenge of class imbalance within \coolname{}, WPD+ is incorporated as a data augmentation step. It includes two tunable parameters: the number of sampled frames from the original dataset and the inclusion of the synthetic dataset. To determine the optimal configuration, we conducted two ablation experiments.  Fig.~\ref{subfig:wpd+_frames} shows that sampling 6 frames results in the optimal performance, while increasing the number of frames beyond this point leads to performance degradation. Fusing too many frames can introduce noise and redundant information that may overwhelm the model, for instance, excessive fusion might result in overlapping objects and distorted boundaries when too many new objects are pasted into the scene. Furthermore, Fig.~\ref{subfig:synthetic_wpd+} illustrates that the synthetic dataset plays a key-role in refining semantic prediction of top-rare classes. Notably, using the synthetic dataset is both efficient and practical, as it allows us to customize sensor configurations to align with the target dataset and define specific objects within the scene for downstream applications without incurring any labor cost.
\begin{figure}[h]
\vspace{-2mm}
\centering
    \includesvg[width=0.92\columnwidth]{pics/post_processing.svg}
    \label{fig:post_processing}
    \vspace{-1mm}
\caption{Ablation on various post-processing approaches: KNN~\cite{2019rangenet++}, NLA~\cite{zhao2021fidnet}, and NNRI are first compared in a single-range setting. *NNRI denotes the multi-range variant of our post-processing method. For fair evaluation, we also extend KNN to operate on multiple range images (*KNN). Dotted lines indicate the metrics for 2D predictions.}
\vspace{-1mm}
% Our WPD+ includes sampling 6 frames from the same dataset and 2 from the synthetic one.
\label{fig:post_processing}
\end{figure}
\PAR{Post-Processing} We explore the performance of various post-processing techniques on both efficacy and efficiency in Fig.~\ref{fig:post_processing}. Regarding conventional KNN~\cite{2019rangenet++} as the baseline, NLA~\cite{zhao2021fidnet} demonstrates similar performance in both accuracy and latency. In contrast, we deploy our approach (NNRI) in the standard setting as well and observe a significant improvement: inference time is cut nearly 16\% compared to KNN, while mAcc and mIoU increase by 2.8\% and 2\%, respectively. Unlike KNN, NNRI avoids the computational cost of Gaussian kernel calculations for distance weighting and directly performs nearest neighbor searches on the range image instead of in 3D space, which further reduces computational overhead. NNRI interpolates class-wise scores based on relative depths rather than directly voting on hard labels, relying more on weighted information from nearest neighbors, which is the major reason why it outperform other post-processing approaches.\\
Next, switching to \coolname{} mode, we first implement an extension of KNN, which iteratively gathers votes from all points in each sub-cloud and aggregates them for the final prediction. While this extension improves the accuracy, it comes at approximately doubled latency cost. Conversely, when NNRI is adapted to all sub-clouds, it consistently provides notable improvements in both efficacy and efficiency. As a reference, we included evaluation scores on 2D predictions, showing that \coolname{} with NNRI significantly narrows the accuracy gap between 2D and 3D predictions. This suggests that our approach effectively mitigates the "many-to-one" problem and offers substantial gains in segmentation performance.
\begin{figure}[t]
\vspace{-1mm}
  \centering
  % First image (subfigure a)
  \begin{subfigure}[b]{0.15\textwidth}
  \centering
    \includegraphics[width=\textwidth]{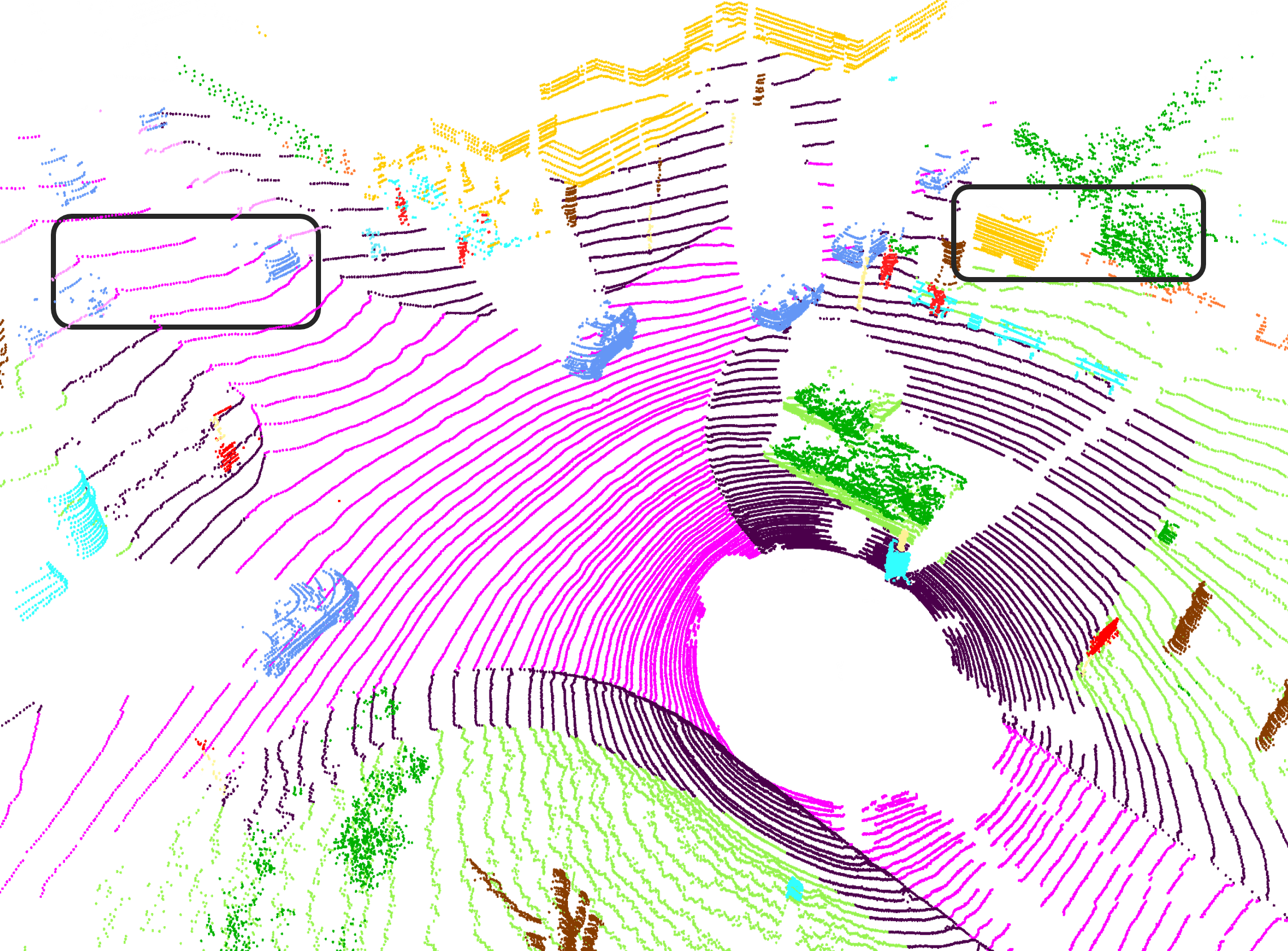}
    \caption{Scene}
  \end{subfigure}
  % Second image (subfigure b)
  \begin{subfigure}[b]{0.1\textwidth}
  \centering
    \includegraphics[width=\textwidth]{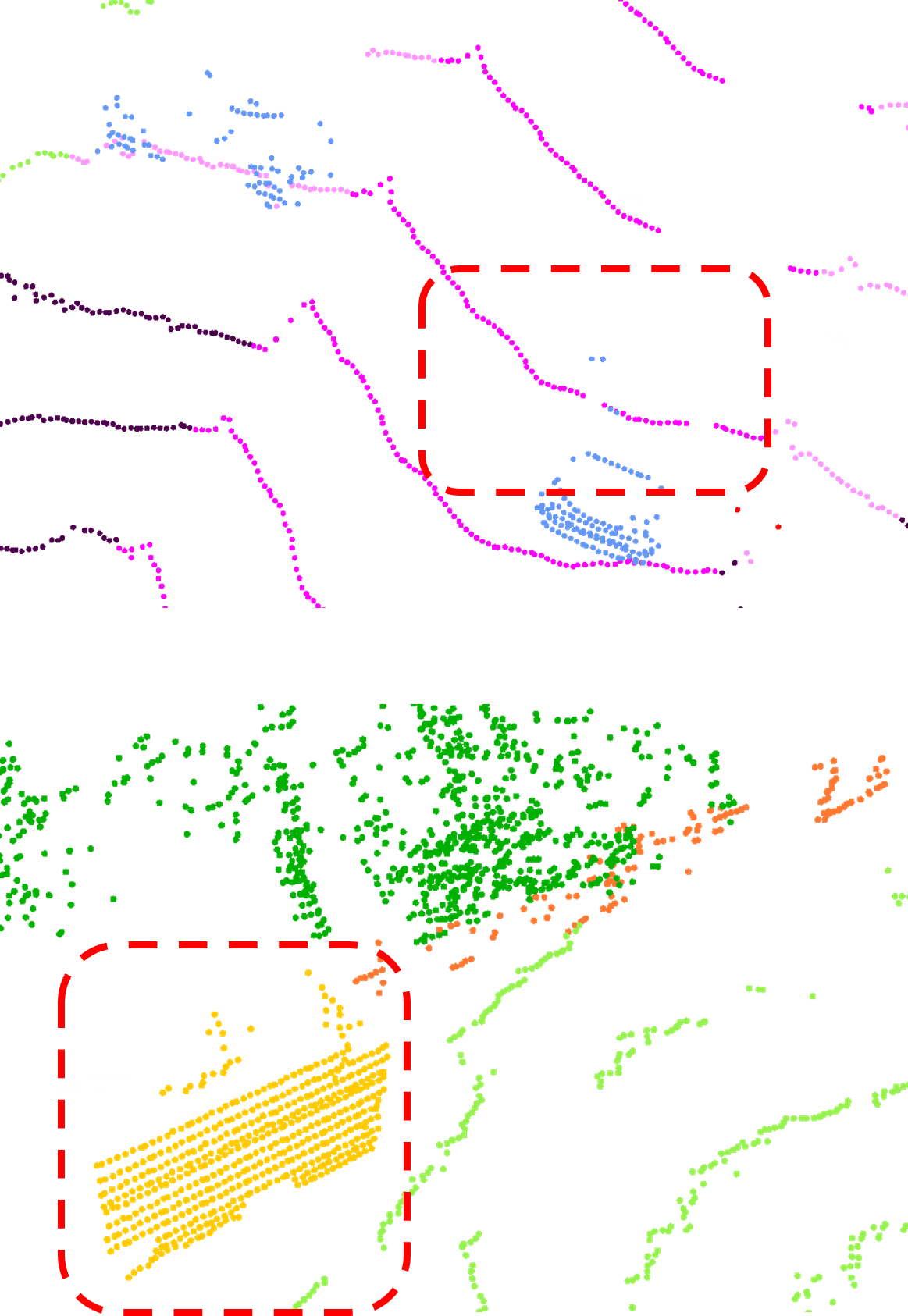}
    \caption{GT}
  \end{subfigure}
  % Third image (subfigure c)
  \begin{subfigure}[b]{0.1\textwidth}
  \centering
    \includegraphics[width=\textwidth]{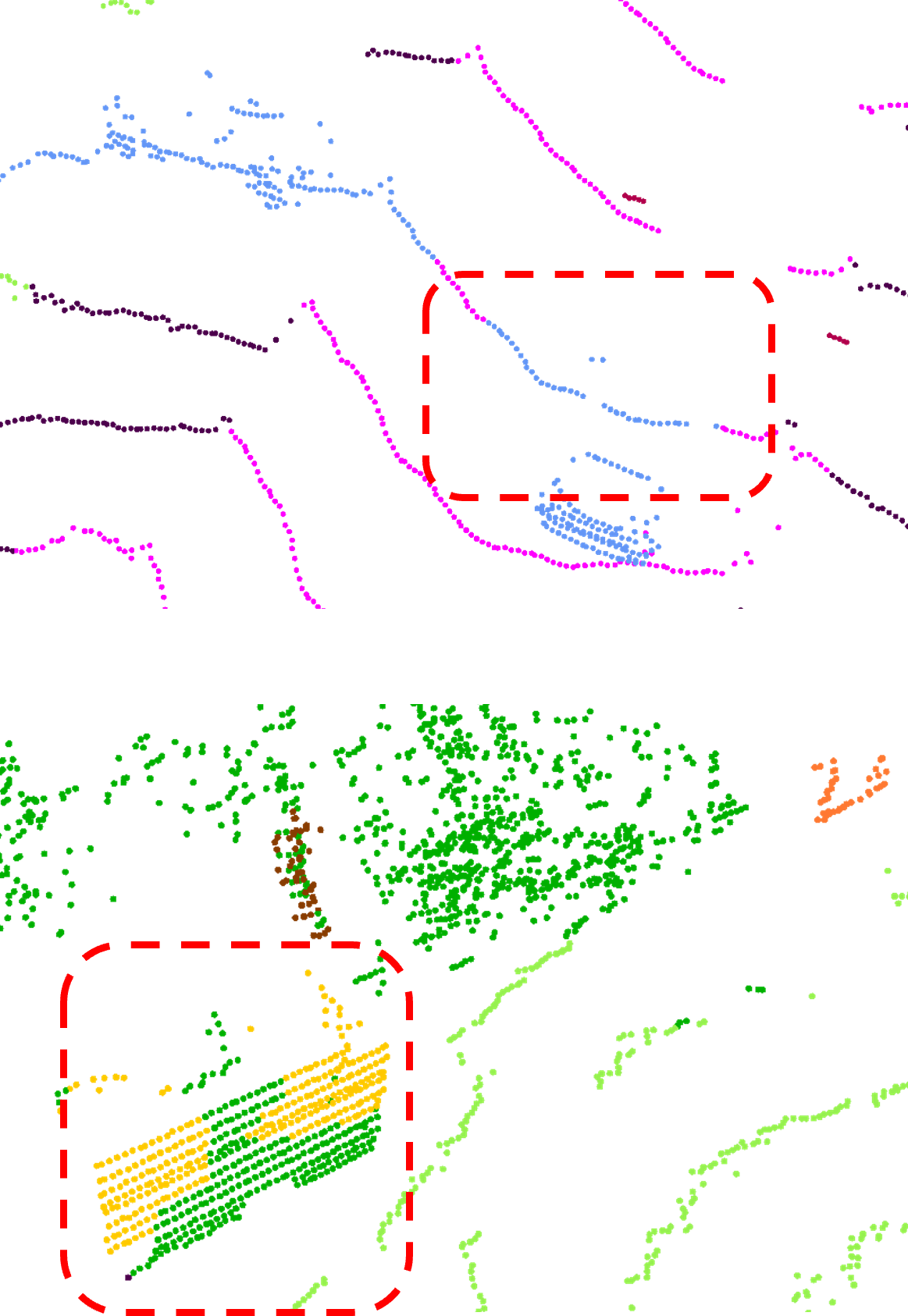}
    \caption{Constant}
  \end{subfigure}
  \begin{subfigure}[b]{0.1\textwidth}
  \centering
    \includegraphics[width=\textwidth]{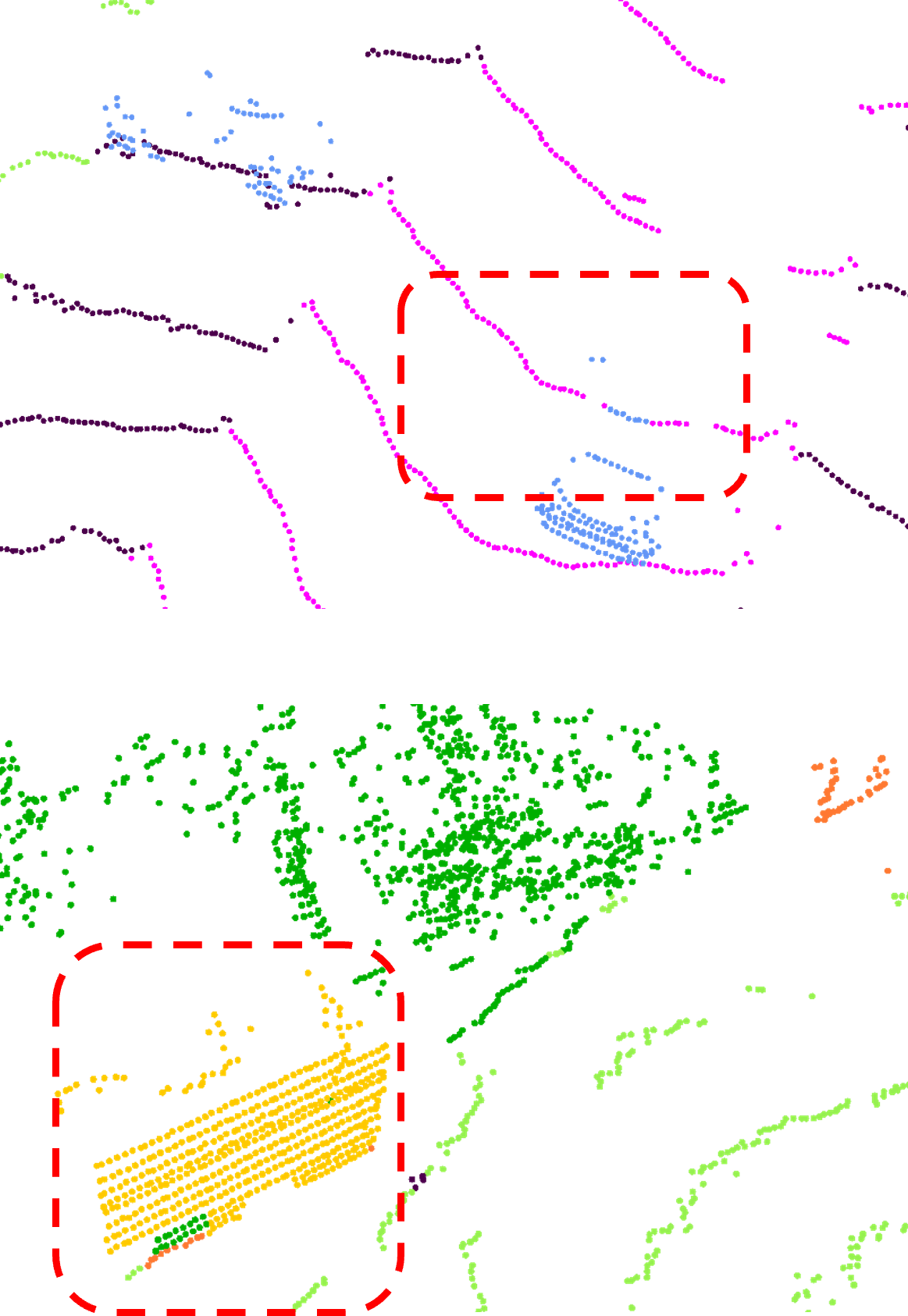}
    \caption{Adaptive}
  \end{subfigure}
  \caption{Segmentation results with different cut-off values in NNRI: in the case of constant value (set at 1), overlapping points of \textcolor{SemanticKITTI-Road}{Road} are partially misclassified as \textcolor{SemanticKITTI-Car}{Car} in the top image. Similarly, in the bottom image, half of the points that belong to \textcolor{SemanticKITTI-Building}{Building} are incorrectly predicted as \textcolor{SemanticKITTI-Veg}{Vegetation}.}
  \label{fig:cutoff-values}
  \vspace{-1mm}
\end{figure}
Furthermore, we optimize the implementation of NNRI by leveraging an adaptive cut-off value to filter valid nearest neighbors. This parameter is derived by approximating the internal LiDAR geometry and the distance-density relationship of 3D points. To illustrate its impact, qualitative results are provided in Fig.~\ref{fig:cutoff-values}. As shown, the adaptive cut-off refines semantic predictions by better accommodating objects with varying density scales.
% As introduced in Sec.~\ref{sec:post-processing}, a simple LiDAR model is built to approximate the distance-density function of 3D points and compute cut-off values for valid neighbors extraction. To verify the necessity of the step, some qualitative results are provided in Fig.~\ref{fig:cutoff-values}. As can be seen, the adaptive cut-off values refine semantic predictions by better accommodating objects with varying density scales. 
\begin{figure}[h]
\vspace{-2mm}
\centering
\begin{subfigure}{.235\textwidth}
\centering
    \includesvg[width=\columnwidth]{pics/number_of_subclouds.svg}
    \caption{}
    \label{subfig:number_of_subclouds}
\end{subfigure}
\begin{subfigure}{.235\textwidth}
\centering
    \includesvg[width=\columnwidth]{pics/resolution.svg}
    \caption{}
    \label{subfig:resolution}
\end{subfigure}
\caption{(a) Ablation on number of sub-clouds ($N$) with fixed image resolution of $64\times512$. (b) Ablation on image width $W$ ($N$ is fine-tuned for each configuration, respectively).}
% Our WPD+ includes sampling 6 frames from the same dataset and 2 from the synthetic one.
\vspace{-2mm}
\label{fig:ablation_postprocessing}
\end{figure}
\PAR{Input Configuration} In Fig.~\ref{subfig:number_of_subclouds}, we investigate in the effect on the number of splitting. It shows that increasing the number of sub-clouds beyond 4 degrades performance due to insufficient occupancy and projection artifacts. Conversely, fewer sub-clouds speeds up the inference but compromises accuracy. Overall, our choice of 3 sub-clouds optimally balances efficiency and accuracy. In Fig~\ref{subfig:resolution}, we further test various azimuth resolution of input images. The experimental results indicate that width of $512$ delivers the best mIoU scores. Increasing or decreasing the azimuth resolution beyond this point causes a slight performance drop. % Nevertheless, all tested configurations outperform the baseline (the first row in Tab.\ref{tab:ablation_ts}), demonstrating the superb effectiveness of our approach on range-view LiDAR semantic segmentation.
%
% \begin{figure}[!]
%   \centering
%     \includegraphics[width=\linewidth]{}
%     \caption{Evaluation of different input resolutions and number of sub-clouds on SemanticKITTI~\cite{behley2019semantickitti} \textit{val} set.}
%     \label{}
% \end{figure}
% %

% \PAR{Test Time Augmentation}

% \PAR{Pretraining}
% if this is not working, just put the results in the supplementary material

% \subsection{Qualitative Results}

% \subsection{Limitations}

%% file: tables/qualitative_results.tex
\begin{figure}[t]
    \centering
    \hspace*{-6mm}
    \begin{tabular*}{\linewidth}{m{-4cm}m{2.3cm}m{2.3cm}m{2.3cm}}
        % Row 1
    \quad & \hspace{3mm} \fontsize{8}{8}\selectfont Groundtruth & \hspace{6mm} \fontsize{8}{8}\selectfont Baseline & \hspace{7mm} \fontsize{8}{8}\selectfont \coolname{}\\
    \rotatebox{90}{\fontsize{7}{10}\selectfont SalsaNext~\cite{cortinhal2020salsanext}}&\includegraphics[width=0.16\textwidth]{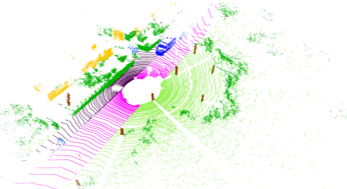}&\includegraphics[width=0.16\textwidth]{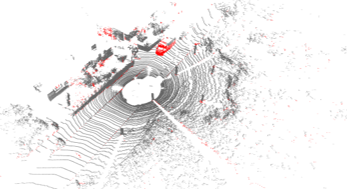}&\includegraphics[width=0.16\textwidth]{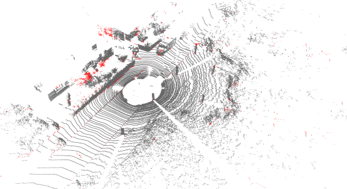}\\  
    \rotatebox{90}{\fontsize{7}{10}\selectfont FIDNet~\cite{zhao2021fidnet}}&\includegraphics[width=0.15\textwidth]{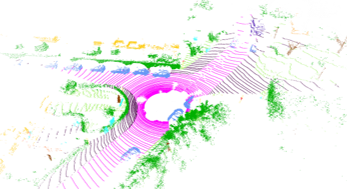}&\includegraphics[width=0.15\textwidth]{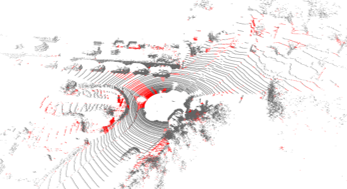}&\includegraphics[width=0.15\textwidth]{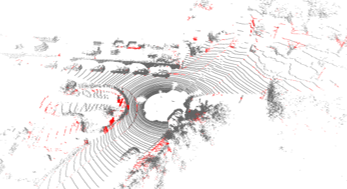}\\
    \rotatebox{90}{\fontsize{7}{10}\selectfont CENet~\cite{cheng2022cenet}}&\includegraphics[width=0.16\textwidth]{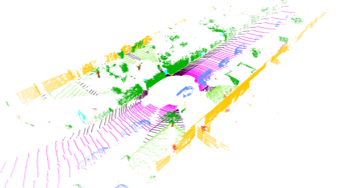}&\includegraphics[width=0.16\textwidth]{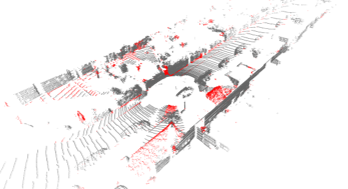}&\includegraphics[width=0.16\textwidth]{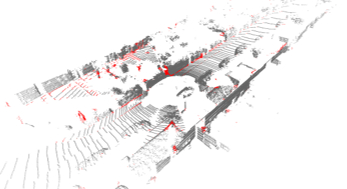}\\  
    \rotatebox{90}{\fontsize{7}{10}\selectfont RangeViT~\cite{ando2023rangevit}}&\includegraphics[width=0.16\textwidth]{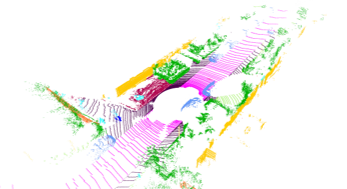}&\includegraphics[width=0.16\textwidth]{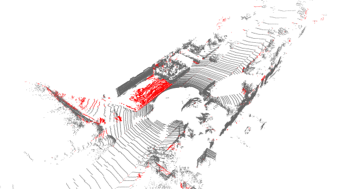}&\includegraphics[width=0.16\textwidth]{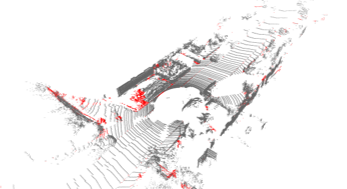}\\ 
    \end{tabular*}
    
    \vspace{-0.3mm}
    \caption{\textbf{Qualitative results on SemanticKITTI}\cite{behley2019semantickitti} Points in \textcolor{incorrect}{red} and \textcolor{correct}{gray} represent incorrect and correct predictions, respectively. $^\star$More examples are provided in the supplementary material.}
    \vspace{-2mm}
    \label{fig:qualitative_results}
\end{figure}

%% file: tables/semantickitti_test_v2.tex
\begin{table*}[t]

\vspace{-3mm}
\centering\scalebox{0.68}{
\begin{tabular}{c|c|cccccccccccccccccccc}
\toprule
\multicolumn{21}{c}{~\textbf{SemanticKITTI \textit{test} set}}
\\\midrule
 \textbf{Method}  & \rotatebox{0}{$\text{mIoU}$} & \rotatebox{0}{car} & \rotatebox{0}{bicy} & \rotatebox{0}{moto} & \rotatebox{0}{truc} & \rotatebox{0}{o.veh} & \rotatebox{0}{ped} & \rotatebox{0}{b.list} & \rotatebox{0}{m.list} & \rotatebox{0}{road} & \rotatebox{0}{park} & \rotatebox{0}{walk} & \rotatebox{0}{o.gro} & \rotatebox{0}{build} & \rotatebox{0}{fenc} & \rotatebox{0}{veg} & \rotatebox{0}{trun} & \rotatebox{0}{terr} & \rotatebox{0}{pole} & \rotatebox{0}{sign}
\\\midrule
SalsaNext~\cite{cortinhal2020salsanext} &  $59.5$ & $91.9$ & $48.3$ & $38.6$ & $38.9$ & $31.9$ & $60.2$ & $59.0$ & $\mathbf{19.4}$ & $91.7$ & $63.7$ & $75.8$ & $\mathbf{29.1}$ & $90.2$ & $64.2$ & \underline{$81.8$} & $63.6$ & $\mathbf{66.5}$ & $54.3$ & $62.1$
\\
\textcolor{orange}{\text{\faBolt}}\textbf{SalsaNext}\textcolor{orange}{\text{\faBolt}}  &   \underline{$63.3$} &    \underline{$94.7$} &   \underline{$52.9$} &   \underline{$55.7$} &     \underline{$57.3$} &   \underline{$50.2$} &   \underline{$65.5$} &   \underline{$70.9$} &   \underline{$13.0$} &   \underline{$92.6$} &   $\mathbf{69.0}$ &   \underline{$77.7$} &   \underline{$20.5$} &   \underline{$90.4$} &   \underline{$65.8$} &   $80.8$ &   \underline{$65.0$} &   $63.4$ &   \underline{$55.4$} &   \underline{$62.4$}
\\
$^\diamond$\textcolor{orange}{\text{\faBolt}}\textbf{SalsaNext}\textcolor{orange}{\text{\faBolt}}  &   $\mathbf{64.8}$ &    $\mathbf{95.1}$ &   $\mathbf{55.5}$ &   $\mathbf{56.5}$ &     $\mathbf{60.1}$ &   $\mathbf{53.7}$ &   $\mathbf{69.6}$ &   $\mathbf{74.1}$ &   $11.4$ &   $\mathbf{93.0}$ &   \underline{$68.9$} &   $\mathbf{78.9}$ &   $20.4$ &   $\mathbf{91.1}$ &   $\mathbf{67.6}$ &   $\mathbf{82.0}$ &   $\mathbf{66.7}$ &   \underline{$65.0$} &   $\mathbf{58.1}$ &   $\mathbf{64.1}$
\\
\arrayrulecolor[rgb]{0.4,0.4,0.4} \hline
$^\dagger$FIDNet~\cite{zhao2021fidnet}  &  $51.3$ & $90.4$ & $28.6$ & $30.9$ & $34.3$ & $27.0$ & $43.9$ & $48.9$ & \underline{$16.8$} & $90.1$ & $58.7$ & $71.4$ & $19.9$ & $84.2$ & $51.2$ & $78.2$ & $51.9$ & $64.5$ & $32.7$ & $50.3$
\\
FIDNet &  $59.5$ & $93.9$ & \underline{$54.7$} & $48.9$ & $27.6$ & $23.9$ & $62.3$ & $59.8$ & $\mathbf{23.7}$ & $90.6$ & $59.1$ & $75.8$ & $26.7$ & $88.9$ & $60.5$ & \underline{$84.5$} & $64.4$ & \underline{$69.0$} & $53.3$ & $62.8$
\\
\textcolor{orange}{\text{\faBolt}}\textbf{FIDNet}\textcolor{orange}{\text{\faBolt}}  &  \underline{$65.1$} &    \underline{$95.3$} &   $51.0$ &   \underline{$57.0$} &     \underline{$54.8$} &   \underline{$58.1$} &   \underline{$68.1$} &   \underline{$68.9$} &   $14.4$ &   \underline{$92.3$} &   \underline{$68.3$} &   \underline{$78.0$} &   \underline{$32.3$} &   \underline{$91.6$} &   \underline{$67.6$} &   $83.7$ &   \underline{$66.6$} &   $68.8$ &   \underline{$55.1$} &   \underline{$64.8$}
\\
$^\diamond$\textcolor{orange}{\text{\faBolt}}\textbf{FIDNet}\textcolor{orange}{\text{\faBolt}}  &  $\mathbf{67.4}$ &    $\mathbf{95.8}$ &   $\mathbf{56.7}$ &   $\mathbf{60.7}$ &     $\mathbf{58.1}$ &   $\mathbf{60.3}$ &   $\mathbf{72.5}$ &   $\mathbf{72.9}$ &   $15.8$ &   $\mathbf{93.2}$ &   $\mathbf{69.2}$ &   $\mathbf{79.9}$ &   $\mathbf{34.2}$ &   $\mathbf{91.9}$ &   $\mathbf{69.0}$ &   $\mathbf{84.6}$ &   $\mathbf{68.7}$ &   $\mathbf{70.3}$ &   $\mathbf{59.9}$ &   $\mathbf{66.9}$
\\
\arrayrulecolor[rgb]{0.4,0.4,0.4} \hline
$^{\diamond\dagger}$CENet~\cite{cheng2022cenet}  &  $60.7$ & $92.1$ & $45.4$ & $42.9$ & $43.9$ & $46.8$ & $56.4$ & $63.8$ & \underline{$29.7$} & $91.3$ & $66.0$ & $75.3$ & $31.1$ & $88.9$ & $60.4$ & $81.9$ & $60.5$ & $67.6$ & $49.5$ & $59.1$
\\
$^\diamond$CENet  & $64.7$ & $91.9$ &   \underline{$58.6$} & $50.3$ & $40.6$ & $42.3$ &   {$68.9$} & $65.9$ & $\mathbf{43.5}$ & $90.3$ & $60.9$ & $75.1$ & $31.5$ & $91.0$ & $66.2$ & \underline{$84.5$} & $69.7$ & $\mathbf{70.0}$ &   \underline{$61.5$} &   $67.6$
\\
\textcolor{orange}{\text{\faBolt}}\textbf{CENet}\textcolor{orange}{\text{\faBolt}}  &    \underline{$66.6$} &     \underline{$95.6$} &     $58.5$ &     \underline{$61.6$} &   \underline{$51.7$} &     \underline{$50.2$} &     \underline{$74.5$} &     \underline{$72.4$} &     $23.2$ &   \underline{$91.4$} &     \underline{$69.6$} &     \underline{$77.1$} &   \underline{$31.7$} &     \underline{$91.1$} &     \underline{$66.6$} &     $83.8$ &     \underline{$69.9$} &     $68.3$ &     $60.3$ &   \underline{$68.7$}
\\
$^\diamond$\textcolor{orange}{\text{\faBolt}}\textbf{CENet}\textcolor{orange}{\text{\faBolt}}  &    $\mathbf{68.0}$ &     $\mathbf{95.9}$ &     $\mathbf{61.1}$ &     $\mathbf{62.1}$ &   $\mathbf{57.2}$ &     $\mathbf{59.0}$ &     $\mathbf{77.2}$ &     $\mathbf{74.2}$ &     $12.2$ &   $\mathbf{92.2}$ &     $\mathbf{69.9}$ &     $\mathbf{78.7}$ &   $\mathbf{32.9}$ &     $\mathbf{91.8}$ &     $\mathbf{68.8}$ &     $\mathbf{84.7}$ &     $\mathbf{71.3}$ &     \underline{$69.9$} &     $\mathbf{62.9}$ &   $\mathbf{70.3}$
\\
\arrayrulecolor[rgb]{0.4,0.4,0.4} \hline
% \arrayrulecolor{gray}\hline
RangeViT~\cite{ando2023rangevit} & $64.0$ & $95.4$ & $55.8$ & $43.5$ & $29.8$ & $42.1$ & $63.9$ & $58.2$ & $\mathbf{38.1}$ &   $\mathbf{93.1}$ & $70.2$ &   $\mathbf{80.0}$ &   $32.5$ &   $\mathbf{92.0}$ &   $\mathbf{69.0}$ & $\mathbf{85.3}$ &   $\mathbf{70.6}$ &   $\mathbf{71.2}$ & $\mathbf{60.8}$ & $64.7$
\\
\textcolor{orange}{\text{\faBolt}}\textbf{RangeViT}\textcolor{orange}{\text{\faBolt}}  &   $\mathbf{66.1}$ &   $\mathbf{95.6}$ &   $\mathbf{56.3}$ &   $\mathbf{60.5}$ &   $\mathbf{52.4}$ &   $\mathbf{57.1}$ &   $\mathbf{72.0}$ &   $\mathbf{69.7}$ &   $16.0$ &     $91.6$ &   $\mathbf{71.1}$ &   ${77.3}$ &     $\mathbf{32.7}$ &   $91.4$ &   $67.4$ &   $83.1$ &   $68.0$ &   $68.1$ &   $58.0$ &     $\mathbf{67.5}$
\\\midrule
\end{tabular}}
\bigskip
\centering\scalebox{0.75}{
\begin{tabular}{c|c|cccccccccccccccc}
\multicolumn{18}{c}{~\textbf{nuScenes \textit{val} set}}
\\\midrule
 \textbf{Method~\small{(year)}}  & \rotatebox{0}{$\text{mIoU}$} & \rotatebox{0}{barrier} & \rotatebox{0}{bicy} & \rotatebox{0}{bus} & \rotatebox{0}{car} & \rotatebox{0}{const} & \rotatebox{0}{moto} & \rotatebox{0}{ped} & \rotatebox{0}{traffic.c} & \rotatebox{0}{trailer} & \rotatebox{0}{truck} & \rotatebox{0}{driv} & \rotatebox{0}{o.flat} & \rotatebox{0}{side} & \rotatebox{0}{terrain} & \rotatebox{0}{manm} & \rotatebox{0}{veg}
\\\midrule
SalsaNext &  $72.2$ & $74.8$ & $34.1$ & $85.9$ & $88.4$ & $42.2$ & $72.4$ & $72.2$ & $\mathbf{63.1}$ & $61.3$ & $76.5$ & $96.0$ & $\mathbf{70.8}$ & $71.2$ & $71.5$ & $86.7$ & $84.4$ 
\\
\textcolor{orange}{\text{\faBolt}}\textbf{SalsaNext}\textcolor{orange}{\text{\faBolt}}  &   $\mathbf{74.5}$ &    $\mathbf{75.0}$ &   $\mathbf{34.6}$ &   $\mathbf{90.4}$ &     $\mathbf{90.0}$ &   $\mathbf{43.8}$ &   $\mathbf{79.4}$ &   $\mathbf{72.9}$ &   $58.8$ &   $\mathbf{65.8}$ &   $\mathbf{79.9}$ &   $\mathbf{96.5}$ &   $70.1$ &   $\mathbf{74.0}$ &   $\mathbf{73.9}$ &   $\mathbf{87.6}$ &   $\mathbf{85.6}$
\\
\arrayrulecolor[rgb]{0.8,0.8,0.8} \hline
FIDNet &  $72.7$ & $73.0$ & $36.0$ & $87.8$ & $86.0$ & $45.6$ & $74.1$ & $73.9$ & $62.5$ & $67.1$ & $77.7$ & $94.3$ & $69.8$ & $72.2$ & $72.1$ & $86.1$ & $84.5$ 
\\
\textcolor{orange}{\text{\faBolt}}\textbf{FIDNet}\textcolor{orange}{\text{\faBolt}}  &  $\mathbf{76.6}$ &    $\mathbf{77.6}$ &   $\mathbf{43.5}$ &   $\mathbf{92.9}$ &     $\mathbf{88.1}$ &   $\mathbf{56.5}$ &   $\mathbf{79.5}$ &   $\mathbf{77.7}$ &   $\mathbf{65.3}$ &   $\mathbf{67.0}$ &   $\mathbf{83.1}$ &   $\mathbf{96.6}$ &   $\mathbf{72.8}$ &   $\mathbf{75.0}$ &   $\mathbf{74.5}$ &   $\mathbf{88.5}$ &   $\mathbf{86.8}$ 
\\
\arrayrulecolor[rgb]{0.8,0.8,0.8} \hline
CENet  & $73.7$ & $73.6$ &   $32.9$ & $92.7$ & $87.1$ & $\mathbf{53.5}$ &   $76.1$ & $69.0$ & $58.7$ & $66.8$ & $81.6$ & $95.6$ & $71.1$ & $73.7$ & $73.2$ & $87.5$ & $85.7$
\\
\textcolor{orange}{\text{\faBolt}}\textbf{CENet}\textcolor{orange}{\text{\faBolt}}  &  $\mathbf{76.8}$ &    $\mathbf{76.7}$ &   $\mathbf{45.2}$ &   $\mathbf{93.5}$ &     $\mathbf{90.3}$ &   $49.6$ &   $\mathbf{83.1}$ &   $\mathbf{78.1}$ &   $\mathbf{66.4}$ &   $\mathbf{69.0}$ &   $\mathbf{82.5}$ &   $\mathbf{96.6}$ &   $\mathbf{73.9}$ &   $\mathbf{75.1}$ &   $\mathbf{74.6}$ &   $\mathbf{88.3}$ &   $\mathbf{86.3}$ 
\\
\arrayrulecolor[rgb]{0.8,0.8,0.8} \hline
% \arrayrulecolor{gray}\hline
RangeViT& $75.2$ & $75.5$ & $\mathbf{40.7}$ & $88.3$ & $90.1$ & $49.3$ & $79.3$ & $77.2$ & $\mathbf{66.3}$ &   $65.2$ & $80.0$ &   {$96.4$} &   {$71.4$} &   {$73.8$} &   {$73.8$} & $\mathbf{89.9}$ &   $\mathbf{87.2}$ 
\\
\textcolor{orange}{\text{\faBolt}}\textbf{RangeViT}\textcolor{orange}{\text{\faBolt}}  &   $\mathbf{77.0}$ &   $\mathbf{76.7}$  &   $39.2$ &   $\mathbf{93.0}$ &   $\mathbf{92.0}$ &   $\mathbf{55.2}$ &   $\mathbf{81.6}$ &   $\mathbf{77.2}$ &   $64.9$ &     $\mathbf{70.9}$ &   $\mathbf{84.1}$ &   $\mathbf{96.8}$ &     $\mathbf{74.1}$ &   $\mathbf{75.6}$ &   $\mathbf{75.1}$ &   $88.6$ &   $86.7$
\\\bottomrule
\end{tabular}}
\vspace{-2mm}
\caption{Comparisons of state-of-the-art LiDAR semantic segmentation methods on the \textit{test} set of SemanticKITTI~\cite{behley2019semantickitti} and \textit{val} set of nuScenes~\cite{caesar2020nuscenes} in standard and ~\coolname{} mode. IoU scores are reported in percentages (\%). For each method block, \textbf{bold} and \underline{underline} indicate the \textbf{best} and \underline{second best} result in the column. $^\dagger$Baseline results trained on low-resolution ($64\times512$) range images. $^\diamond$Models inferred with test-time augmentation~\cite{kong2023rethinking}. Note that we did not use model ensembling to further boost the model performance.} 
\vspace{-2mm}
\label{table:semantickitti_nuscenes}
\end{table*}

%% file: tables/val_test.tex
\begin{table}[h]
% \footnotesize
\vspace{-1.5mm}
\renewcommand \arraystretch{1}
\resizebox{\columnwidth}{!}{%
    \begin{tabular}{ r|c|c|c|c|c|c} 
    \toprule
     \textbf{Method~\small{(year)}} & Size & Lat. & Modality& $\oplus$  & $\boxplus$ & $\boxdot$
    \\\midrule
    \textcolor{orange}{\text{\faBolt}}\textbf{SalsaNext}\textcolor{orange}{\text{\faBolt}}  & 6.7M& \textbf{29}  & Range & \textbf{74.5} &  \textbf{64.8} & \underline{64.8}\\
    PolarNet~\cite{zhang2020polarnet}~\small{['20]} & 13.6M & 71 & Polar & 71.0 & 54.9 &57.5 \\
SPVNAS~\cite{tang2020searching}~\small{['20]} & 12.5M & 259 & Voxel & - & \underline{64.7} & \textbf{66.4} \\
    % 3D-MiniNet~\cite{tang2020searching}~\small{['20]} & 4.0M & \underline{33 ms} & Multi-View & - & - & 56.6 \\
RandLA-Net~\cite{xu2020squeezesegv3}~\small{['20]} & 1.2M & \underline{55} & Point & - & - & 53.9 \\
    Tornado-Net~\cite{gerdzhev2021tornado}~\small{['20]} & - & - & Multiple & - & 64.5 & 63.1 \\
    \arrayrulecolor[rgb]{0.4,0.4,0.4} \hline
    \textcolor{orange}{\text{\faBolt}}\textbf{FIDNet}\textcolor{orange}{\text{\faBolt}}  & 6.1M& \textbf{26}  & Range &  \underline{76.6} &  65.6 & \underline{67.4}  \\
     Cylinder3D~\cite{zhu2021cylindrical}~\small{['21]} &  56.3M & 170 &  Voxel & 76.1 & \underline{67.8} & 65.9 \\
     RPVnet~\cite{xu2021rpvnet}~\small{['21]} &  24.8M & 168&  Multiple & \textbf{77.6} & \textbf{68.2} & \textbf{70.3} \\
     FPS-Net~\cite{xiao2021fps}~\small{['21]} &  55.7M & \underline{48} &  Range & - & 54.9 & 57.1 \\
     Lite-HDSeg~\cite{razani2021lite}~\small{['21]} &  - & 50 &  Range & - & 64.4 & 63.8 \\
    \arrayrulecolor[rgb]{0.4,0.4,0.4} \hline
    \textcolor{orange}{\text{\faBolt}}\textbf{CENet}\textcolor{orange}{\text{\faBolt}}  & 6.8M& \textbf{24} & Range & 76.8 &  67.5 &  68.0 \\
 Meta-RSeg~\cite{wang2022meta}~\small{['22]} &  6.8M & \underline{46} & Range & - & 60.3 &61.0\\ 
    PVKD~\cite{pvkd}~\small{['22]} &  14.1M & 76 & Voxel  & 76.0  & 66.4  & 71.2\\
     PTv2~\cite{wu2022ptv2}~\small{['22]} &  12.8M & 213& Point & \textbf{80.2} & \textbf{70.3} & \textbf{72.6}\\
    2DPASS~\cite{yan20222dpass}~\small{['22]} &  26.5M & 119& Multiple & \underline{79.4} & \underline{69.3} & \underline{72.2}\\
     GFNet~\cite{qiu2022gfnet}~\small{['22]} &  - & 100& Multiple & 76.8 & 63.2 & 65.4\\
     \arrayrulecolor[rgb]{0.4,0.4,0.4} \hline
     % FRNet~\cite{xu2023frnet}~\small{['24]}  & 41 & Frustum & 79.0 & 68.7 & 73.3 \\ 
     WaffleIron~\cite{puy2023waffleiron}~\small{['23]}& 6.8M & 143 & Point & 79.1 & 68.0&70.8  \\
     SphereFormer~\cite{lai2023spherical}~\small{['23]} & 32.3M & 165 & Multi & 78.4 & 67.8 &74.8 \\
     PTv3~\cite{wu2024ptv3}~\small{['24]} & 46.2M & 67 & Point & 80.4 &70.8 &74.2 
    \\\bottomrule
    \end{tabular}
    }
    \vspace{-0.2mm}
    \caption{Comparisons of state-of-the-art LiDAR semantic segmentation methods in accuracy (mIoU [\%]) and efficiency (Latency [ms]). All methods are categorized by year of publication. $\oplus$ represents \textit{val} set of nuScenes~\cite{caesar2020nuscenes}, while $\boxplus$ and $\boxdot$ stand for \textit{val} and \textit{test} set of SemanticKITTI~\cite{behley2019semantickitti}. Latency with integration of \coolname{} is measured by processing all sub-clouds.}
    \vspace{-4mm}
    \label{tab:SOTA}
\end{table}

%% file: tables/full_ablation.tex
\begin{table}
\vspace{-1.5mm}
\centering
% \small
\renewcommand{\arraystretch}{1}
\label{tab:supervised}
\resizebox{\linewidth}{!}{
\begin{tabular}{c|c|c|c|c?c|c}
% \boldline
\hline
  STR~\cite{kong2023rethinking} & \coolname{} & WPD+ & MCF & NNRI & \textbf{mIoU} & \textbf{Lat.}\\
\hline
    &  &  & & &63.1 & 44 ms \\ 
\checkmark&  & & & & 62.6& 41 ms\\ 
 & \checkmark & & & & 64.2 & 46 ms\\ 
& \checkmark & \checkmark& & & 65.5 & -\\ 
&  \checkmark&\checkmark &\checkmark & & 65.9 & -\\ 
&  \checkmark&\checkmark &\checkmark & \checkmark& 67.5& 24 ms\\ 

% \coolname{} w/o Mix & 78.1 & 71.8 & \\
% sup. w/ DMB & 77.8 {\small \textcolor{teal}{(+0.1)}} & 71.8 {\small \textcolor{teal}{(+1.4)}}& 73.4 {\small \textcolor{teal}{(+1.3)}}\\
% \arrayrulecolor{gray}\hline
% \coolname{} w/o DMB & 77.2 & 73.0 & 74.1 \\
% \boldline
\hline
\end{tabular}}
\caption{Full ablation study on SemanticKITTI~\cite{behley2019semantickitti} dataset. The first row denotes the baseline results trained in standard mode with high-resolution range images ($64\times2048$). For STR~\cite{kong2023rethinking}, we split the full range image into 4 sub-images, each has the resolution of $64\times512$. For all experiments without NNRI, we employ the conventional KNN~\cite{2019rangenet++} post-processing. }
\label{tab:ablation_full}
\vspace{-2mm}
\end{table}

%% file: sec/5_conclusion.tex
\section{Conclusion}
% In this work, we proposed \coolname{}, a new pipeline to train and infer range-view-based network more efficiently. Compounded with that, we also introduced WPD+ and MCF, two additional data augmentation techniques to increase the model scalability. At last, a novel unsupervised post-processing method was proposed to effectively and rapidly mitigate the accuracy drop led by the "many-to-one" problem. Our approach has presented significant performance improvement across different network architectures.
In this work, we introduce \coolname{}, an optimized multi-range training paradigm designed for range-view LiDAR semantic segmentation. Our framework seamlessly integrates into any range-view-based network. We also develop two data augmentation techniques to address the challenges of exacerbated class imbalance and amplified projection noise within \coolname{}. Additionally, we propose a novel post-processing method tailored to multi-range settings to effectively tackle the "many-to-one" issue. Our approach yields significant improvements in both accuracy and efficiency over baselines across various network architectures on two widely used LiDAR benchmarks. The \textbf{\textit{limitations}} of our current work are twofold: (1) although our data augmentation is powerful, it remains limited in corner cases of extremely low occurrence of certain classes, motivating further exploration of solutions to class imbalance; and (2) the fixed nature of the splitting step may result in the loss of critical information. In future work, we aim to address these limitations and develop a more robust framework that can be generalized to more challenging scenarios, such as adverse weather conditions or limited annotations.
% Despite the limitation in some corner case of class imbalance in the dataset, our approach shows its overall versatility and effectiveness in advancing LiDAR semantic segmentation.

%% file: sec/X_suppl.tex
% \clearpage

\setcounter{page}{1}
\maketitlesupplementary
\appendix

\section{More Implementation details}
In this section, we provide additional details regarding to the overall workflow, proposed components and configurations during training and inference. 
\PAR{Evaluation Metrics} Following prior works, we assess the performance using Intersection-over-Union (IoU) and Accuracy (Acc) for each class, and calculate the mean Intersection-over-Union (mIoU) and mean Accuracy (mAcc) across all classes.
\subsection{Detailed Overview}
\begin{figure*}[h]
  \centering
    \includegraphics[width=0.9\linewidth]{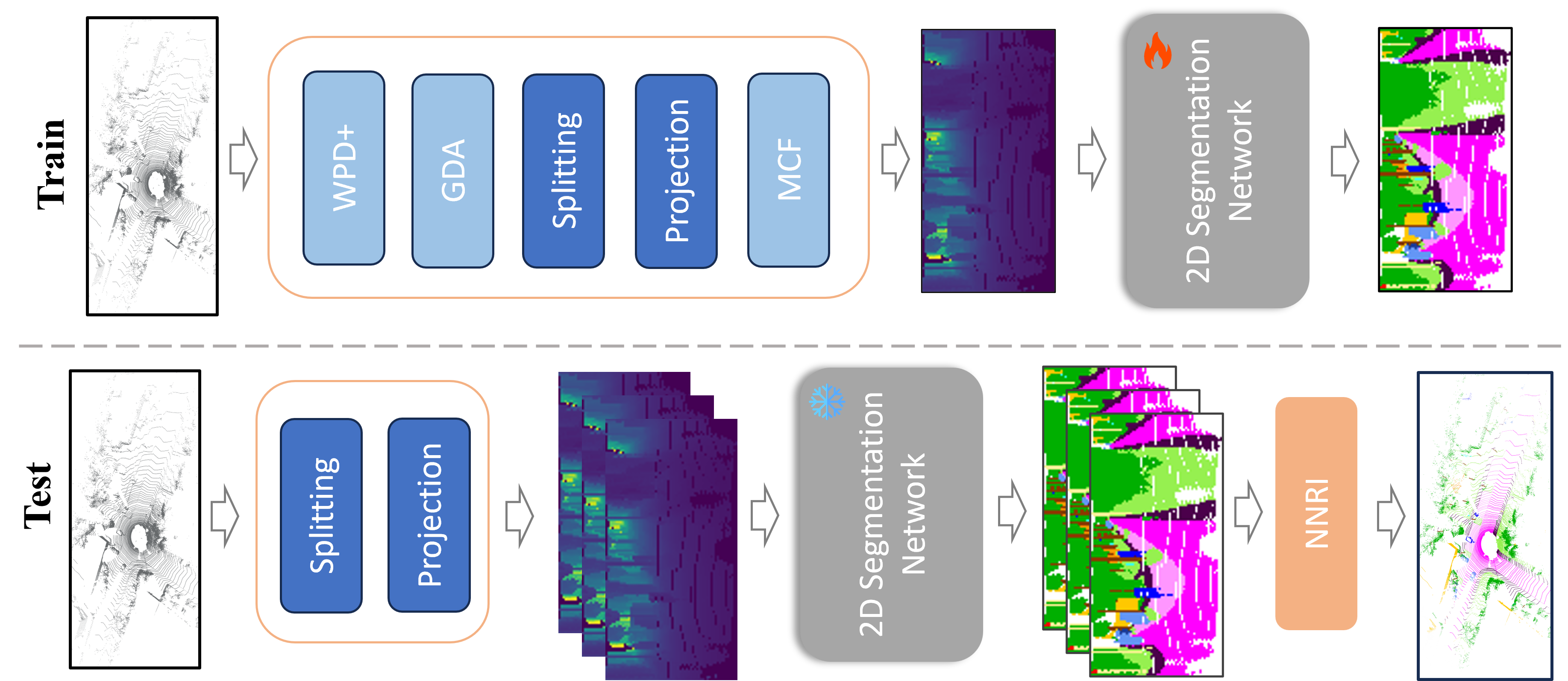}
    \vspace{1mm}
    \caption{Detailed workflow overview: During training, the raw point cloud is pre-processed and augmented to generate the range image. A 2D segmentation network is then trained to predict 2D semantic labels. In testing, all augmentation steps are removed, and range images are stacked as a batch. The 2D predictions are then gathered and processed by NNRI, our proposed post-processing component, to map them into 3D space.}
    \label{fig:supp_detailed_workflow}
\end{figure*}

Each step in the training and testing process is detailed in Fig.~\ref{fig:supp_detailed_workflow}. To maximize the efficiency, the sequence of pre-processing and data augmentation steps is fixed. WPD+ is applied first, before Geometric Data Augmentation (GDA), to avoid performing the same geometric transformation on all sampled frames. Once these two point-level steps are completed, the augmented point cloud is downsampled and divided into multiple sub-clouds, on which projection is performed. Multi-Cloud Fusion (MCF) is applied last, conducting pixel-wise fusion on the range images from each sub-cloud. The same steps are repeated for ground-truth labels to obtain their 2D representation. The final input and ground-truth label are then used to train the 2D semantic segmentation network. During testing, only the pre-processing steps are included, and all range images are fed into the network. Ultimately, we apply NNRI to upsample 2D outputs to gather 3D predictions in an unsupervised manner.
\subsection{Geometric Data Augmentation}
The standard augmentation is utilized commonly in the previous works~\cite{kong2023rethinking, cheng2022cenet, ando2023rangevit}. Hence, we follow the default configurations and include the following geometric transformations in our pipeline:
\begin{itemize}
    \item \PAR{Random Flipping} The point cloud is randomly flipped along the $X$-axis. 
    \item \PAR{Random Translation} The point cloud is randomly jittered in translation. The jittering values range from -5m to 5m, -3m to 3m, and -1m to 0m for $X$-axis, $Y$-axis, and $Z$-axis, respectively.
    \item \PAR{Random Rotation} The point cloud is randomly and globally rotated along yaw, pitch and roll direction. The ranges are set at [-5, 5] in degrees for all axes.
    \item \PAR{Probability} the value of randomness is set at 0.5 for all augmentation components. 
\end{itemize}

\subsection{Runtime Measurement}
In Table 2, we report the inference time of various methods. All measurements are conducted using a single NVIDIA GTX 1080Ti GPU with a fixed batch size of 1. To eliminate the impact of initialization overhead, we discard the first 100 iterations and compute the average runtime over the subsequent 100 iterations.
For methods that could not be measured directly due to GPU limitations, we report the inference times provided in their original papers, most of which were obtained using more powerful GPUs. Notably, all our reported times include data loading, pre-processing, and post-processing steps to ensure a fair comparison.

\subsection{Weighted Paste Drop+}
\begin{figure*}[h]
  \centering
  % First image (subfigure a)
  \begin{subfigure}{0.48\textwidth}
  \centering
    \includegraphics[width=\textwidth]{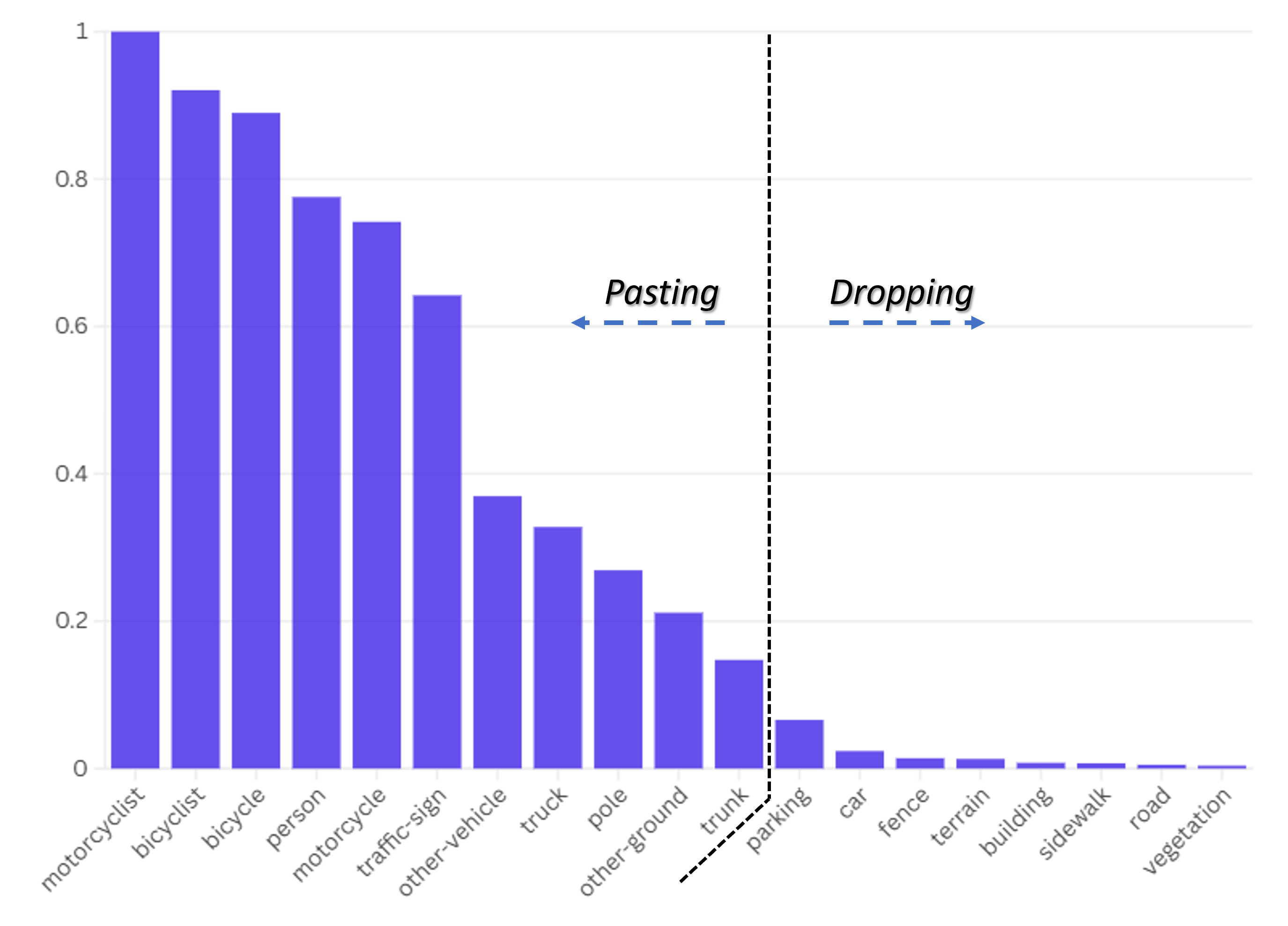}
    \caption{SemanticKITTI~\cite{behley2019semantickitti}}
  \end{subfigure}
  % Second image (subfigure b)
  \begin{subfigure}{0.48\textwidth}
  \centering
    \includegraphics[width=\textwidth]{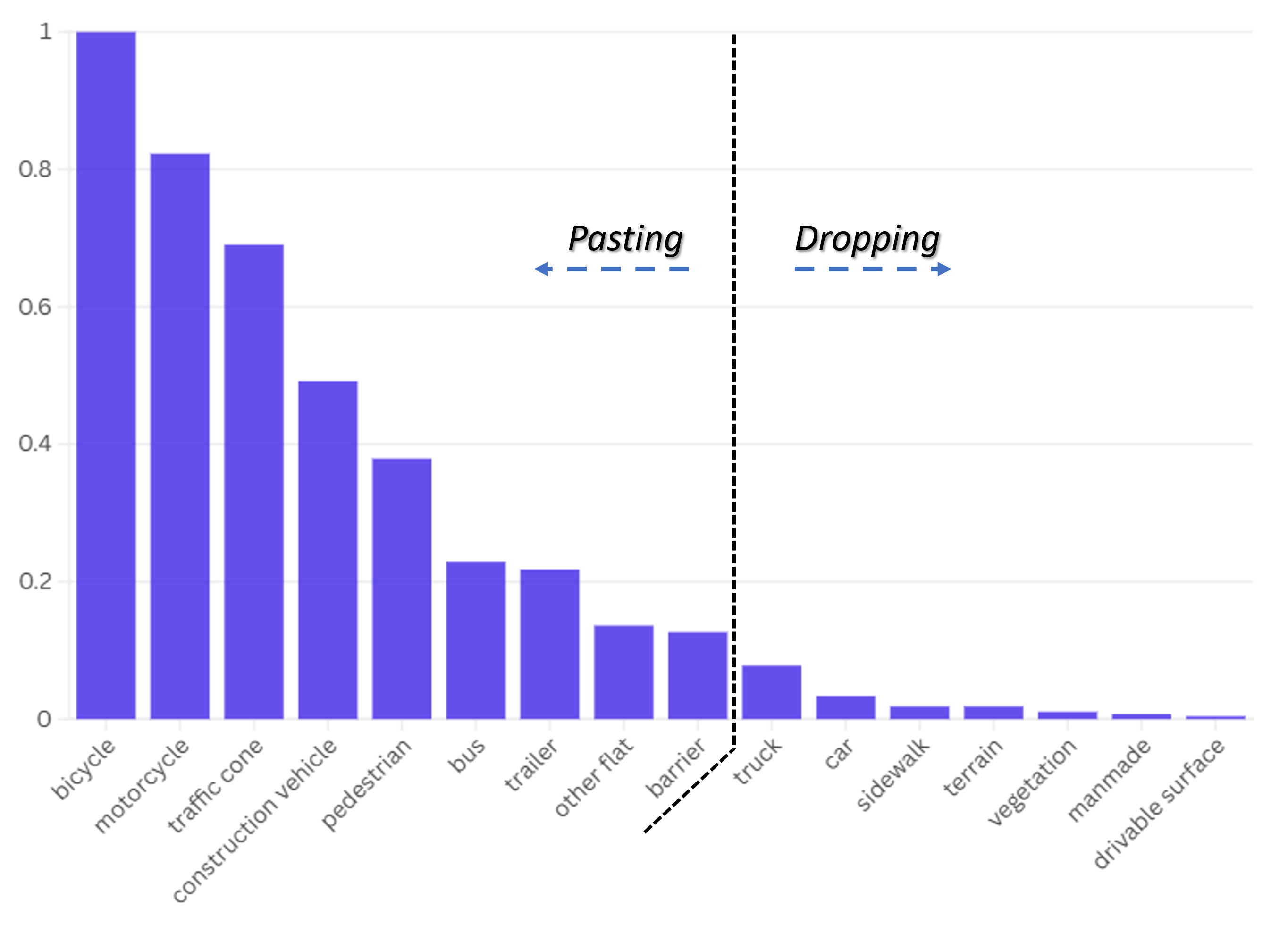}
    \caption{nuScenes~\cite{caesar2020nuscenes}}
  \end{subfigure}
  \caption{Statistical results of class-wise normalized weights on two datasets.}
  \label{fig:supp_statistics}
\end{figure*}

\begin{figure*}[h]
  \centering
    \includegraphics[width=0.9\linewidth]{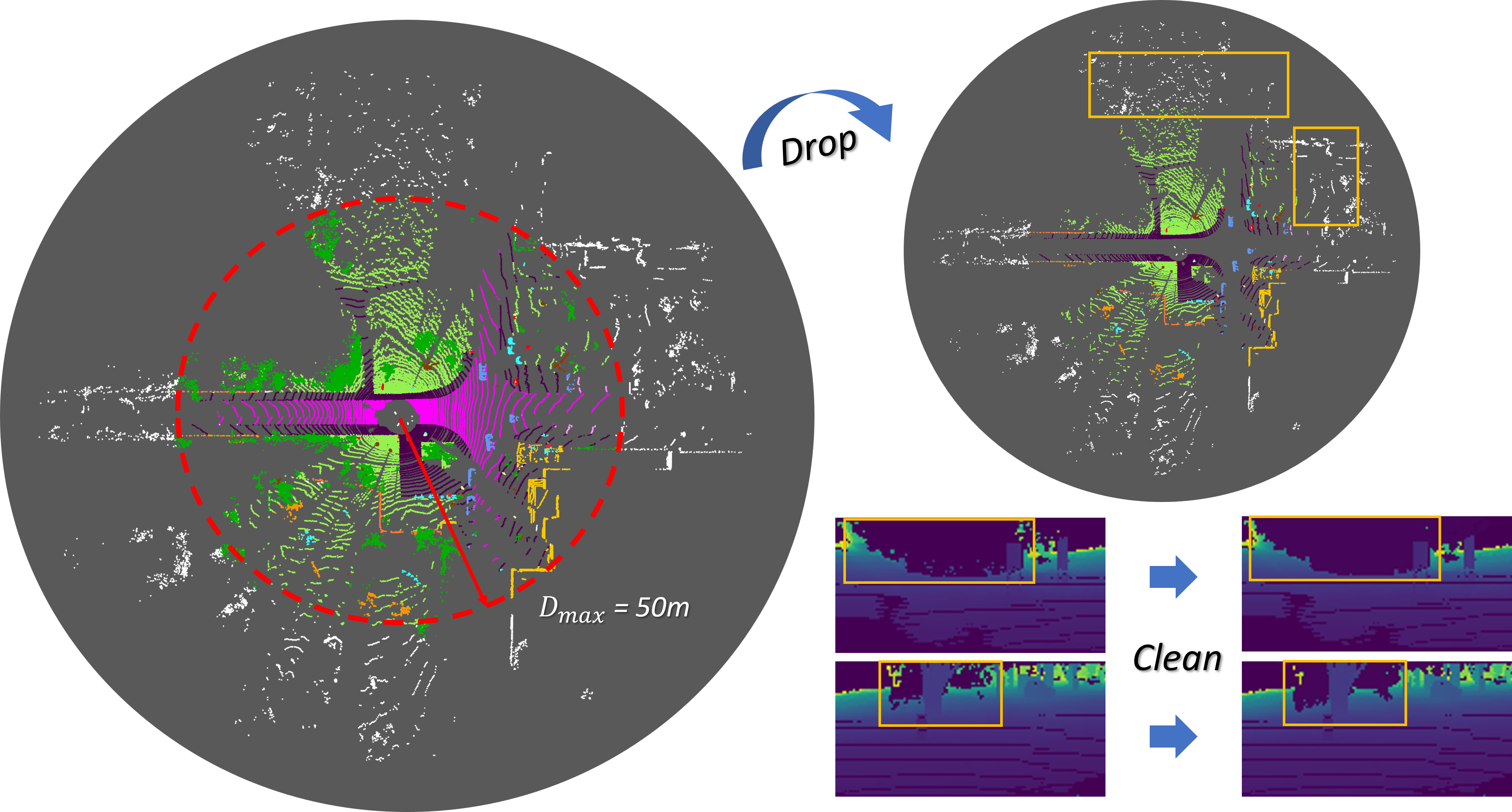}
    \vspace{1mm}
    \caption{An example of dropping points of non-long-tail classes in a point cloud from SemanticKITTI~\cite{behley2019semantickitti} dataset: since the dataset is only annotated within the range of 50 meters, points that are out of this range can result in some artifacts in the projection after points of foreground objects are dropped in the scene. Hence, we insert an additional cleaning step after projection for the dataset.}
    \label{fig:supp_semantic_kitti_artifact}
\end{figure*}

We directly follow the configuration of Weighted Paste Drop (WPD) introduced by \citet{gu2022maskrange}: we first compute the normalized weights from the class-wise frequencies and set the threshold of classifying long-tail and non-long-tail classes at 0.1.  More specifically, we paste the points of classes with weights greater than 0.1 and drop them vice versa. The normalized weight of each class is directly utilized as the probability of pasting and dropping. For reference, we provide the relevant statistics on SemanticKITTI\footnote{\url{https://github.com/PRBonn/semantic-kitti-api}}~\cite{behley2019semantickitti} and nuScenes\footnote{\url{https://www.nuscenes.org/nuscenes}}~\cite{caesar2020nuscenes} in Fig.~\ref{fig:supp_statistics}.
\begin{figure*}[h]
  \centering
    \includegraphics[width=0.9\linewidth]{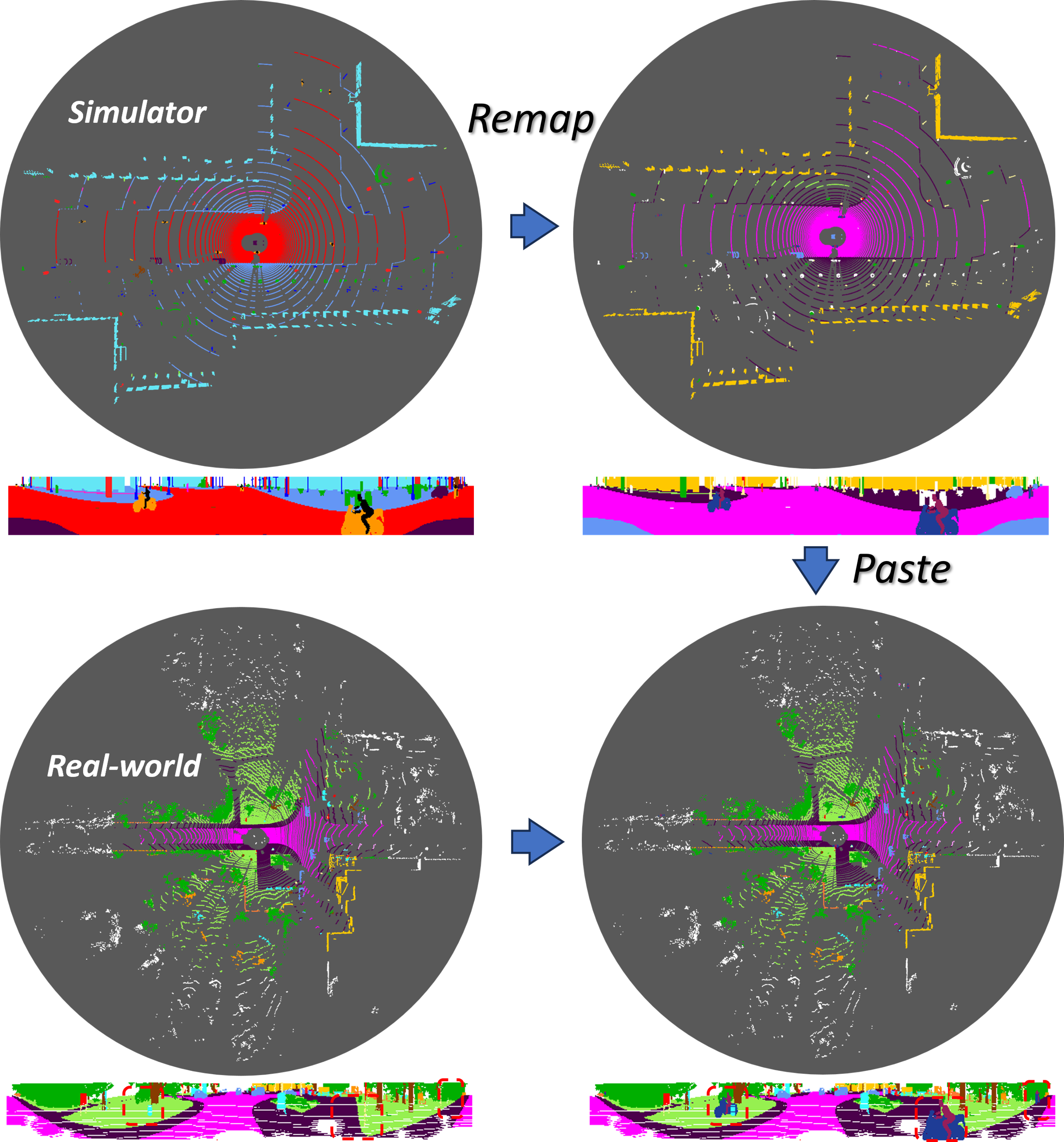}
    \vspace{1mm}
    \caption{An example of pasting points of long-tail classes from synthetic data onto real-world data is shown, with the fused point cloud and range image visualized in the bottom right. Since WPD+ does not account for spatial differences in scene-related contexts, some unrealistic scenarios may arise, such as a motorcyclist riding on a non-drivable surface. Nevertheless, these augmentation steps enhance the semantic richness of the scene. Given that long-tail classes typically correspond to small and dynamic objects, the negative impact of domain differences remains minimal in this case.}
    \label{fig:supp_carla_kittiwpd+}
\end{figure*}
\\
When applying WPD+ to the point clouds in the SemanticKITTI dataset, we observe that the resulting range images sometimes contain artifacts outside of the contextual distributions. As shown in Fig.~\ref{fig:supp_semantic_kitti_artifact}, points at farther distances are left unlabeled in the dataset, as they are less relevant for most perceptual tasks. However, these points can introduce additional noise into the projection when non-long-tail foreground points are excluded. Such outliers can adversely impact training stability, so we further clean the projected range images by filtering out unlabeled representations when training the model with SemanticKITTI dataset.
\subsection{Synthetic Dataset}
In addition to copy-pasting points of long-tail classes from the same dataset, we use Carla Simulator~\cite{dosovitskiy2017carla} to further enhance class balance. This subsection details the process of data collection. \\
Since WPD+ pastes points without spatial transformation and does not account for sensor difference, it is essential that the simulator precisely matches the sensor configuration of the target dataset to prevent introducing out-of-context points during augmentation. We therefore configure the simulator separately for each dataset, based on their respective sensor specifications, as detailed in Tab.~\ref{tab:supp_sensor_configuration}. Next,  we set up the scene-related contexts. Based on statistics shown in Fig.~\ref{fig:supp_statistics}, we initialize the scene with 20 dynamic actors randomly selected from vehicle categories \textit{busses, motorcycles and bicycles} in the Carla Simulator, along with 30 \textit{pedestrians} with random movement. Town10 serves as the static background map for the scene, and we record 2000 frames each for SemanticKITTI and nuScenes.  
\begin{table}[t]
% \vspace{-3mm}
% \footnotesize
\renewcommand \arraystretch{1}
\resizebox{\columnwidth}{!}{%
    \begin{tabular}{ l|c|c|c } 
    \hline
     Dataset& V.FoV [deg] & H.FoV [deg] & Range [m]   \\ 
    \hline
    Sem.KITTI & $-25 \sim 3$ & $-180 \sim 180$ & $2 \sim 50$ \\
    nuScenes & $-30 \sim 10$ & $-180 \sim 180$ & $2 \sim 80$ \\
    \hline
    \end{tabular}
    }
    % \vspace{-0.3mm}
\caption{LiDAR specifications of two datasets.}
    \label{tab:supp_sensor_configuration}
\end{table}\\
After collecting the dataset, we manually define a class mapping scheme between the real and synthetic datasets. The complete mapping function is detailed in Tab.~\ref{tab:supp_mapping}. Generally, positive correspondences can be identified; however, an exception exists where the \textit{rider} class in the synthetic dataset represents both \textit{motorcyclist} and \textit{bicyclist}. To make more effective use of the synthetic dataset, we split and reassign the labels for this class. Since \textit{rider} always appears alongside \textit{motorcycle} and \textit{bicycle} in the scene, we assign new labels by finding the nearest neighbor in 3D space for each \textit{rider} point and categorizing it in a binary manner. We provide an example of the pasting process with the synthetic data in Fig.~\ref{fig:supp_carla_kittiwpd+} for the comprehensive understanding.
\begin{table}[t]
\centering
\small
\begin{tabular}{cc}
\toprule
\textbf{Sem.KITTI} & \textbf{Carla}  \\
\midrule
bicycle  &  bicycle    \\
motorcycle  &  motorcycle     \\
truck  &  truck     \\
other vehicle  &  bus, train     \\
person  &  pedestrian    \\
bicyclist  &  rider*     \\
motorcyclist  &  rider*     \\
other ground  &  ground     \\
trunk  &  -     \\
pole  &  pole     \\
traffic sign  &  traffic sign     \\
\bottomrule
\end{tabular}
\begin{tabular}{cc}
\toprule
\textbf{nuScenes} & \textbf{Carla}  \\
\midrule
barrier  &  -    \\ 
bicycle  &  bicycle    \\ 
bus  &  bus   \\ 
const. vehicle  &  -    \\ 
motorcycle  &  motorcycle    \\ 
pedestrian  &  pedestrian    \\ 
traffic cone  &  -    \\ 
trailer  &  -   \\ 
other flat  &  ground    \\ 
\bottomrule
\end{tabular}
\caption{Class mapping dictionaries: we map only the long-tail classes and omit mappings where there is ambiguity in finding corresponding class names.}
\label{tab:supp_mapping}
\end{table}

\subsection{Multi-Cloud Fusion}
\begin{figure*}[h]
  \centering
  % First image (subfigure a)
  \begin{subfigure}{1\textwidth}
  \centering
    \includegraphics[width=\textwidth]{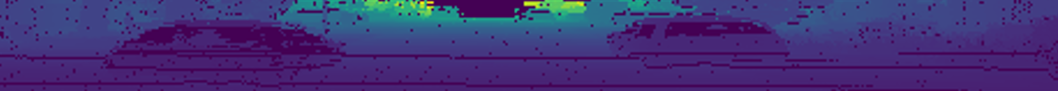}
    \caption{FC, $64\times2048$}
  \end{subfigure}
  % Second image (subfigure b)
  \begin{subfigure}{0.33\textwidth}
  \centering
    \includegraphics[width=\textwidth]{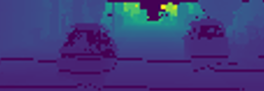}
    \caption{FC, $64\times512$}
  \end{subfigure}
  % Third image (subfigure c)
  \begin{subfigure}{0.33\textwidth}
  \centering
    \includegraphics[width=\textwidth]{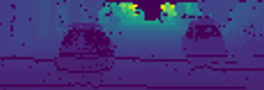}
    \caption{SC, $64\times512$}
  \end{subfigure}
  \begin{subfigure}{0.33\textwidth}
  \centering
    \includegraphics[width=\textwidth]{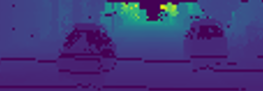}
    \caption{SC, $64\times512$ (MCF)}
  \end{subfigure}
  \caption{Visualization of cropped range images with varying resolutions and configurations of point clouds. Here, FC denotes the full cloud, and SC indicates the sub-cloud. The final image is augmented with Multi-Cloud Fusion (MCF).}
  \label{fig:supp_occupancy}
\end{figure*}

To better illustrate the impact of Multi-Cloud Fusion (MCF) during data augmentation, we provide an example in Fig.~\ref{fig:supp_occupancy}. Notably, range images with high azimuth resolution tend to have a greater number of empty pixels, which are distributed randomly and can distort object geometry within the scene. Reducing the azimuth resolution produces a more complete range image; however, occupancy declines again if only the sub-cloud is available for projection. MCF addresses this issue by filling unoccupied pixels, revisiting data from other sub-clouds at the same 2D positions. In this way, MCF effectively mitigates the problem while preserving the underlying geometry of the scene. 

\subsection{Multi-range KNN}
The KNN-based post-processing approach~\cite{2019rangenet++} has been widely used in prior works. To apply this method with the \coolname{} setup, we extend it to handle multiple range images, allowing us to infer 3D semantic labels for a single full point cloud. The pseudo code is provided in Algo.~\ref{algo:supp_KNN}. Briefly, we gather KNN votes for all points from each image iteratively, accumulating them across images. Each point is then assigned the class with the highest vote count.
\begin{algorithm}[h]
	\SetAlgoLined
    \SetKwInOut{Define}{Define}\SetKwInOut{Input}{Input}\SetKwInOut{Output}{Output}
    \Define {$N = N_{max}$ sub-clouds.\\
            The annotation contains $C$ classes}
    \Input{Range images $R_{ranges}$ with size $N \times H \times W$,\\
           Label images $L_{labels}$ with size $N\times H \times W$,\\
           Arrays $R_{all}(p)$ with range values for all points,\\
           Image coordinates $(u_{all}, v_{all})$ for all points, \\
           }
	\Output{Array $Labels$ with predicted labels for all points.}
	\BlankLine
    % \State \textbf{sajodnakjdnija}
    $Vote(p) = \texttt{Zeros}(p)$\\
    \ForEach{$i$ in $1 : N$}{
        $R_{range} = R_{ranges}(i)$\\ 
        $L_{label} = L_{labels}(i)$\\
        $Vote(p) \mathrel{+}= \mathbf{KNN}(R_{range}, L_{label}, R_{all}, (u_{all}, v_{all}))$  
    }
    $Labels = \texttt{argmax}_{c\in C}(Votes(p))$
    \caption{Multi-range KNN}\label{algo:supp_KNN}
\end{algorithm}

\subsection{Pre-training}
In our main experiments, we test the effectiveness of \coolname{} using four different networks: SalsaNext~\cite{cortinhal2020salsanext}, FIDNet~\cite{zhao2021fidnet}, CENet~\cite{cheng2022cenet}, and RangeViT~\cite{ando2023rangevit}. For specific network designs, we refer readers to the respective original papers. Notably, RangeViT~\cite{ando2023rangevit} demonstrated the advantages of using pretrained models on image datasets~\cite{russakovsky2015imagenet}. Following this approach, we experiment with initializing network weights pre-trained on Cityscapes~\cite{cordts2016cityscapes}. While we apply this pre-training strategy to the other three convolution-based networks, it yields only minor improvements compared to random initialization. We hypothesize that Convolutional Neural Networks (CNNs) are less effective than Vision Transformers (ViTs) in terms of transferability and scalability, as ViTs are better equipped to learn long-range dependencies within the dataset and significantly less biased towards local textures than CNNs~\cite{dosovitskiy2020image, vaswani2017attention}. Consequently, we do not conduct further experiments with other pre-trained weights on the three CNNs.

\section{More quantitative results}
\subsection{Detailed Results}
In Tab.~\ref{tab:supp_semantickitti_full}, we present the class-wise evaluation results of \coolname{}-boosted approaches compared to various LiDAR semantic segmentation methods on SemanticKITTI~\cite{behley2019semantickitti}, including point-wise, voxel-wise, hybrid, and range-wise approaches. In Tab.~\ref{table:supp_nuscenes_testset}, we additionally demonstrate the results from nuScenes~\cite{caesar2020nuscenes} leaderboard. Due to a limited number of entries, we do not categorize methods by publication year and evaluate only a single model using \coolname{}. Clearly, \coolname{} enables the model to achieve superior accuracy compared to most methods and delivers performance comparable to the best-performing approach, all while maintaining significantly higher computational efficiency.\\
Since range-view methods generally offer greater efficiency, it is valuable to compare performance without the use of test-time augmentation. We summarize the results in Tab.~\ref{tab:supp_semantickitti_notta}. Rather than comparing with a broad set of range-view approaches introduced over the past decade, we focus on a selection of the most recent and representative methods, as they demonstrate the best performance among them. Although \coolname{}-boosted networks may not exhibit significantly superior segmentation accuracy compared to the latest approaches, they excel in computational efficiency and usage flexibility. For instance, RangeFormer~\cite{kong2023rethinking} requires high computational resources due to its multiple Vision Transformer blocks~\cite{dosovitskiy2020image}, which contain a substantial number of model parameters. Similarly, TFNet~\cite{li2024tfnet} processes a sequence of LiDAR data rather than a single frame to incorporate temporal features, and its performance is highly dependent on the sequence length.

\subsection{RangeFormer}
To further asses the potential of our method in approaching the state-of-the-art performance, we reproduced the code for RangeFormer~\cite{kong2023rethinking}, the current best-performing range-view method. Results are shown in Tab.~\ref{tab:rangeformer}. Due to the absence of official code, minor discrepancies exist between our reproduced results and the original paper. Nonetheless, integrating our method significantly boosts inference speed while achieving comparable accuracy to SOTA methods of other representations.
\begin{table}[h]
\centering
% \footnotesize
% \vspace{-6.4mm}
\begin{small}
\renewcommand \arraystretch{1}
% \resizebox{\columnwidth}{!}{%
{\fontsize{8}{10}\selectfont
    \begin{tabular}{ c|c|c|c|c}
    \hline
    Model & Params. & S.KITTI \textit{val} & nuScenes \textit{val} &Lat.\\
    \hline
    *RangeFormer &24.3M & 68.1 & 77.1 & -\\
    RangeFormer & - & 67.0 & 76.5 & 87 ms\\
    +\coolname & 24.1M&  68.9 & 78.2 & 55 ms   \\
    \hline
    Cylinder3D~\cite{zhu2021cylindrical} & - & 65.9 & 76.1 & - \\
    SphereFormer~\cite{lai2023spherical} & - & 67.8 & 78.4 & - \\
    PVKD~\cite{pvkd} & - & 66.4 & 76.0 & -\\
    \hline   
    % \coolname-s & {\small $1\times64\times512$}& {\small $1\times64\times512$}  &65.9 & 78.4 & 16 ms
     % texbf{Method} & cell2 & cell3 \\ 
    \end{tabular}
    }
    \vspace{-0.3mm}
    \caption{*: Results reported in the original paper [20].}
    \vspace{-5mm}
    \label{tab:rangeformer}
\end{small}
\end{table}

\subsection{Comparison of Data Augmentation}
We compare our two data augmentation methods with existing approaches that share the same objective. The results are presented in Tables~\ref{tab:pasting} and~\ref{tab:filling}, evaluated on the \textit{val} set of the SemanticKITTI dataset~\cite{behley2019semantickitti}.
\begin{table}[h]
\centering
\vspace{-3mm}
% \footnotesize
% \vspace{-6.4mm}
\begin{small}
\renewcommand \arraystretch{1}
% \resizebox{\columnwidth}{!}{%
{\fontsize{8}{10}\selectfont
    \begin{minipage}{0.45\linewidth}
    \centering
    \begin{tabular}{c|c}
    \hline
    Aug. & mIoU\\
    \hline
    RangePaste~\cite{kong2023rethinking}  &  65.4\\
    % WPD [15]  &  66.1\\
    Mix3D~\cite{nekrasov2021mix3d} & 64.6\\
    Ins. CutMix~\cite{xu2021rpvnet} & 66.2\\
    WPD+ & 67.5 \\
    \hline   
    % \coolname-s & {\small $1\times64\times512$}& {\small $1\times64\times512$}  &65.9 & 78.4 & 16 ms
     % texbf{Method} & cell2 & cell3 \\ 
    \end{tabular}
    
    % \vspace{-0.3mm}
    \caption{Augmentation methods for class imbalance.}
% \caption{Ablation on SemanticKITTI \textit{val} set: Models A and B use standard training with $64\times2048$ input resolution, while C and D use \coolname{} training with $1(3)\times64\times512$ resolution. E-H are ablation on number of sub-images and $64\times512$ resolution is fixed. }
    \vspace{-1.5mm}
    \label{tab:pasting}
    \end{minipage}
    \hfill
    \begin{minipage}{0.45\linewidth}
        \centering
        \begin{tabular}{c|c}
        \hline
        Aug. & mIoU\\
        \hline
        RangeUnion~\cite{kong2023rethinking}  &  66.1\\
        RangeIP~\cite{xu2023frnet} &  66.9\\
        MCF & 67.5 \\
        \hline   
        \end{tabular}
        \caption{Augmentation methods for projection noise.}
        \label{tab:filling}
    \end{minipage}
    }
    \vspace{-3.5mm}
\end{small}
\end{table}

\subsection{Do other standard hyper-parameters matter?}
For consistency, we unify the hyperparameter settings across all experiments. To ensure that improvements in segmentation accuracy are not simply due to changes in hyperparameters, we additionally retrain each network using the default settings specified in their respective original papers. The corresponding results are reported in Table~\ref{tab:baseline_comparison}.
\begin{table}[h]
\centering
% \footnotesize
% \vspace{-6.4mm}
\begin{small}
\renewcommand \arraystretch{1}
% \resizebox{\columnwidth}{!}{%
{\fontsize{8}{10}\selectfont
    \begin{tabular}{ c|cc|cc}
    \hline
    \multirow{2}{*}{Model} & \multicolumn{2}{c|}{Baseline} & \multicolumn{2}{c}{\coolname{}} \\
     & mIoU & Lat. & mIoU & Lat.\\
    \hline
    CENet [5] & 62.6 & 44 ms & 65.7 & 24 ms\\
    FIDNet [49]& 58.9 & 46 ms  & 63.4 & 26 ms\\
    SalsaNext [7]& 59.0  & 51 ms & 64.1&29 ms\\
    \hline   
    % \coolname-s & {\small $1\times64\times512$}& {\small $1\times64\times512$}  &65.9 & 78.4 & 16 ms
     % texbf{Method} & cell2 & cell3 \\ 
    \end{tabular}
    }
    % \vspace{-0.3mm}
    \caption{We retrain the networks with default parameter settings of the original baselines on SemanticKITTI \textit{val} set. These include batch size, optimizer, learning rate, weight decay, training epochs and loss weights. }
% \caption{Ablation on SemanticKITTI \textit{val} set: Models A and B use standard training with $64\times2048$ input resolution, while C and D use \coolname{} training with $1(3)\times64\times512$ resolution. E-H are ablation on number of sub-images and $64\times512$ resolution is fixed. }
    \label{tab:baseline_comparison}
\end{small}
\end{table}

\input{tables/supp_semantickitti.tex}
\input{tables/supp_nuscenes_testset.tex}
\input{tables/supp_semantickitti_notta.tex}
\section{More qualitative results}
\begin{figure*}[h]
  \centering
    \includegraphics[width=1\linewidth]{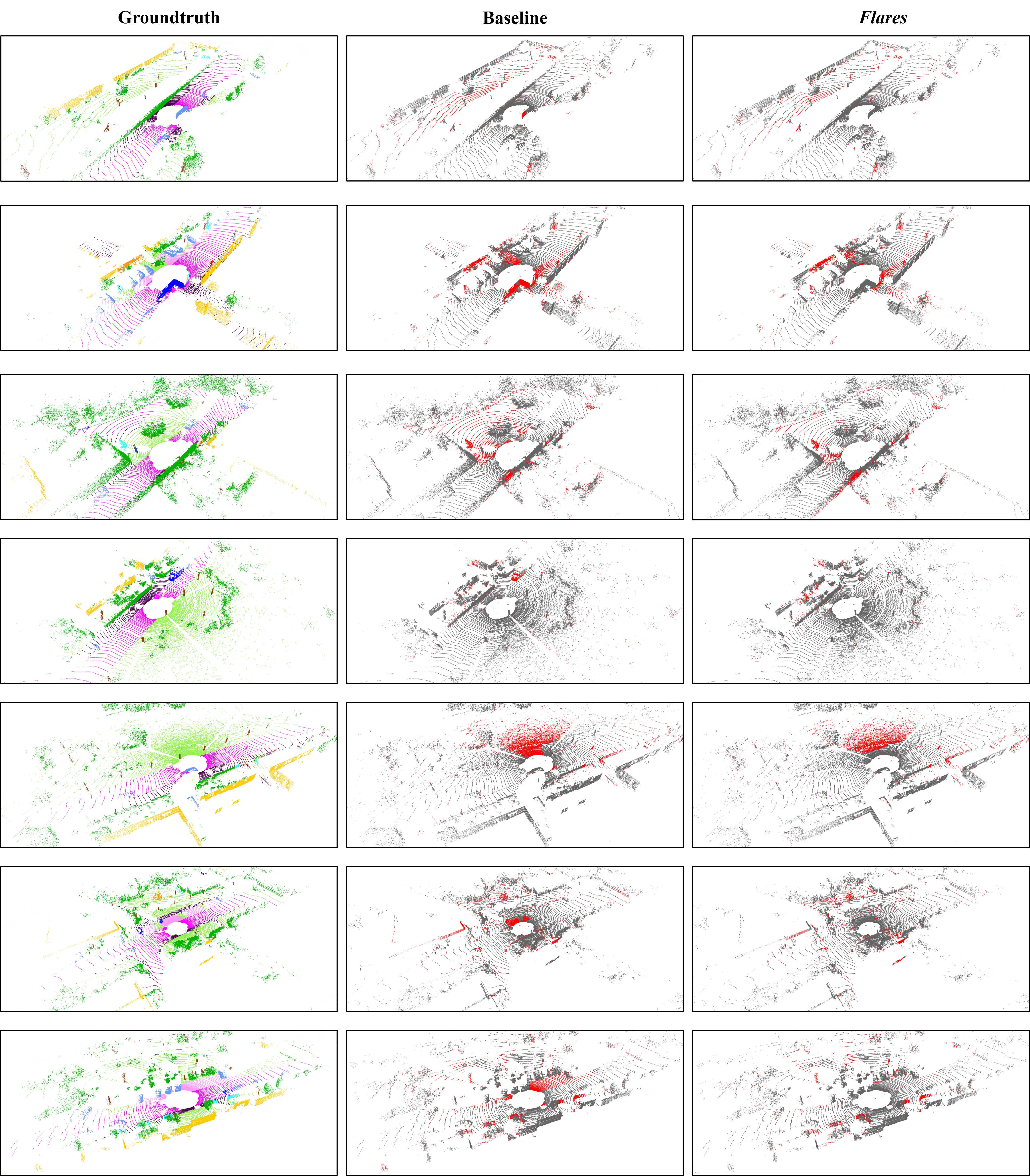}
    \vspace{1mm}
    \caption{Additional qualitative results of \textbf{SalsaNext}~\cite{cortinhal2020salsanext}: points in \textcolor{incorrect}{red} and \textcolor{correct}{gray} represent incorrect and correct predictions, respectively. The groundtruth image is color-coded to differentiate between classes.}
    \label{fig:supp_salsanext_qualitative}
\end{figure*}

\begin{figure*}[h]
  \centering
    \includegraphics[width=1\linewidth]{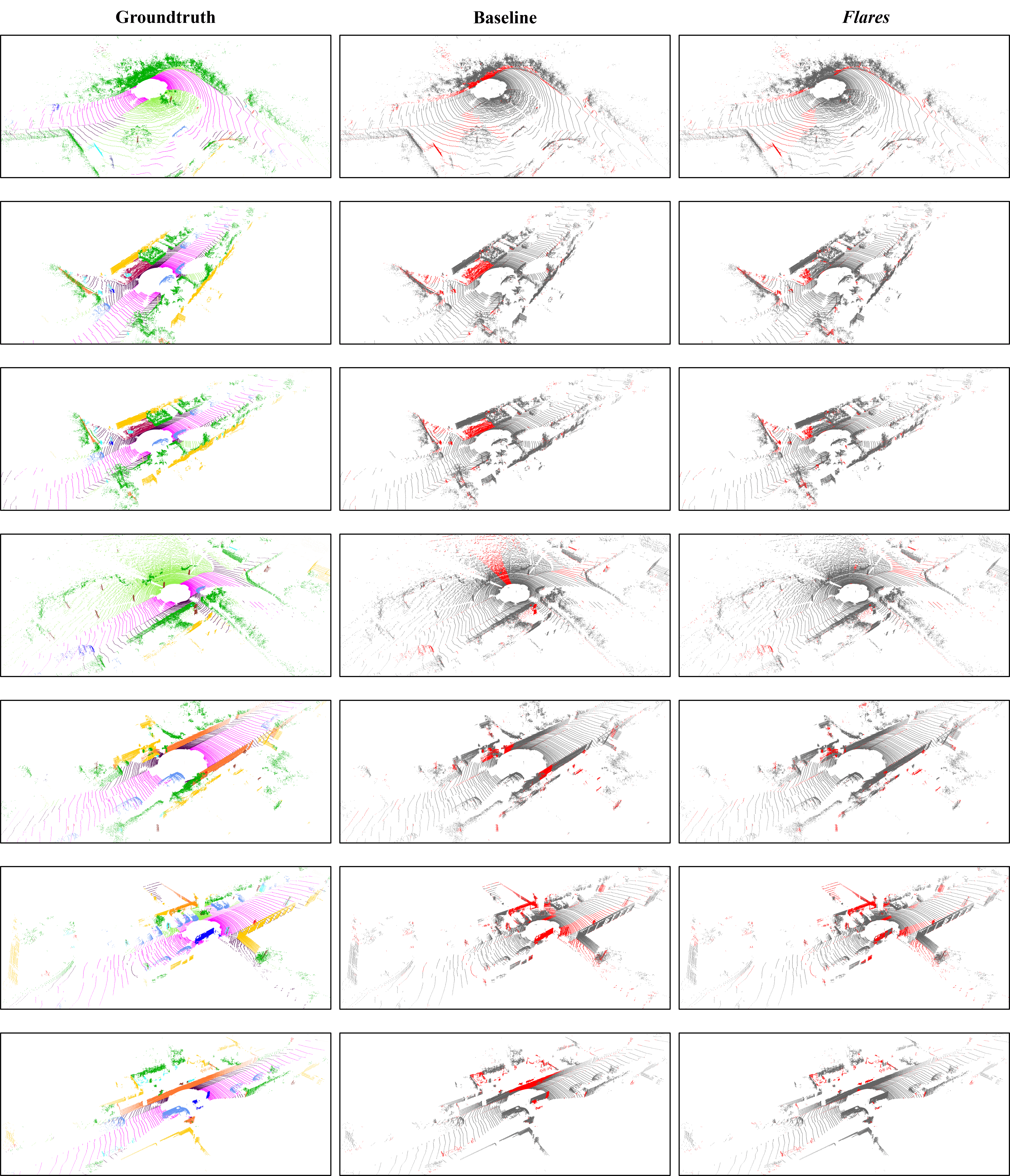}
    \vspace{1mm}
    \caption{Additional qualitative results of \textbf{RangeViT}~\cite{ando2023rangevit}: points in \textcolor{incorrect}{red} and \textcolor{correct}{gray} represent incorrect and correct predictions, respectively. The groundtruth image is color-coded to differentiate between classes.}
    \label{fig:supp_rangevit_qualitative}
\end{figure*}

We provide additional qualitative results of various networks, with and without the integration of \coolname{}, in Fig.~\ref{fig:supp_salsanext_qualitative}$\sim$\ref{fig:supp_rangevit_qualitative}. Compared to the baseline, \coolname{} enhances predictions in scenes with diverse geometries. Notably, segmentation accuracy is significantly improved for foreground and close-to-sensor objects, which are more critical for perception systems than distant ones.
\begin{figure*}[h]
  \centering
    \includegraphics[width=1\linewidth]{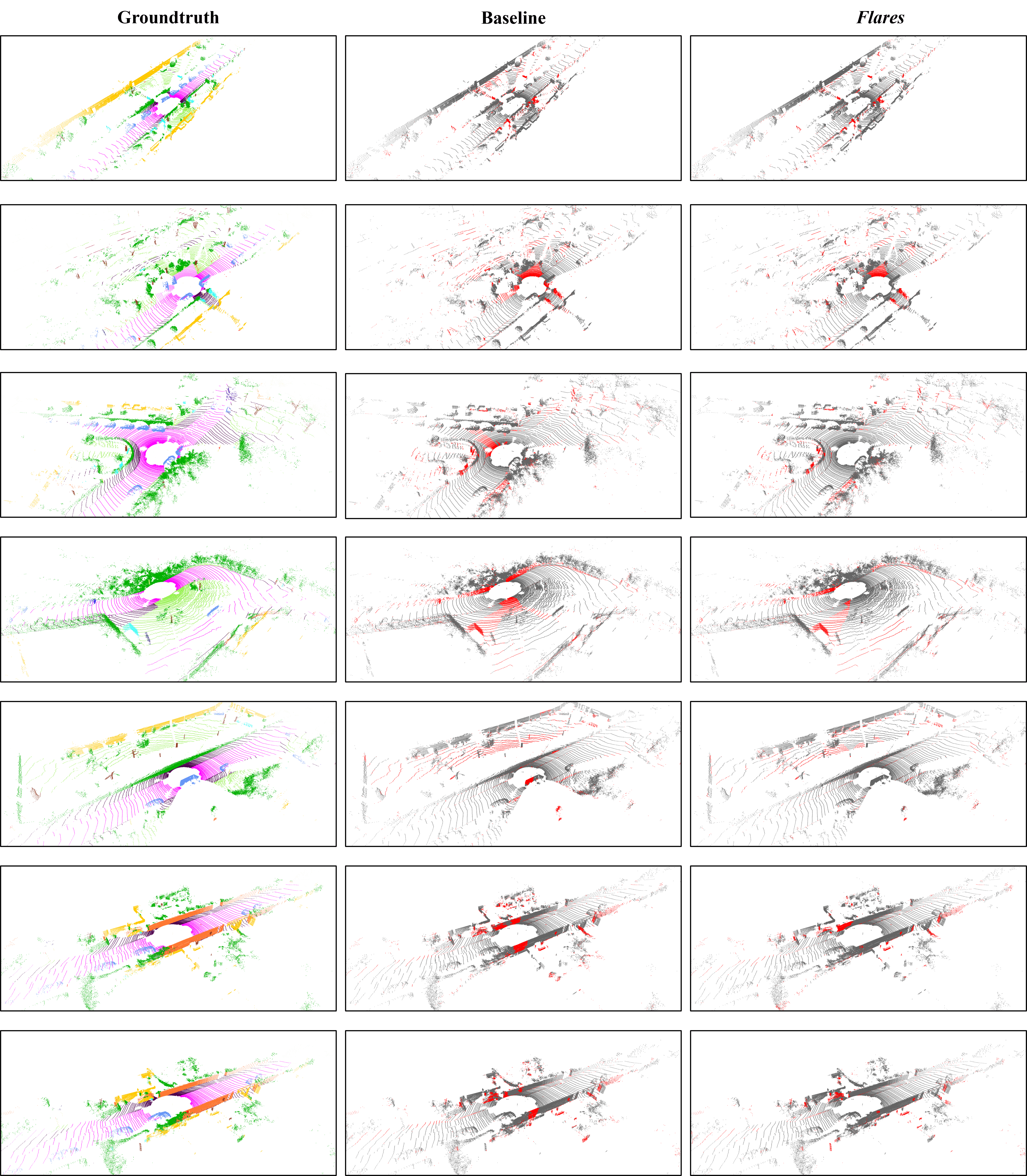}
    \vspace{1mm}
    \caption{Additional qualitative results of \textbf{FIDNet}~\cite{zhao2021fidnet}: points in \textcolor{incorrect}{red} and \textcolor{correct}{gray} represent incorrect and correct predictions, respectively. The groundtruth image is color-coded to differentiate between classes.}
    \label{fig:supp_fidnet_qualitative}
\end{figure*}
\begin{figure*}[h]
  \centering
    \includegraphics[width=1\linewidth]{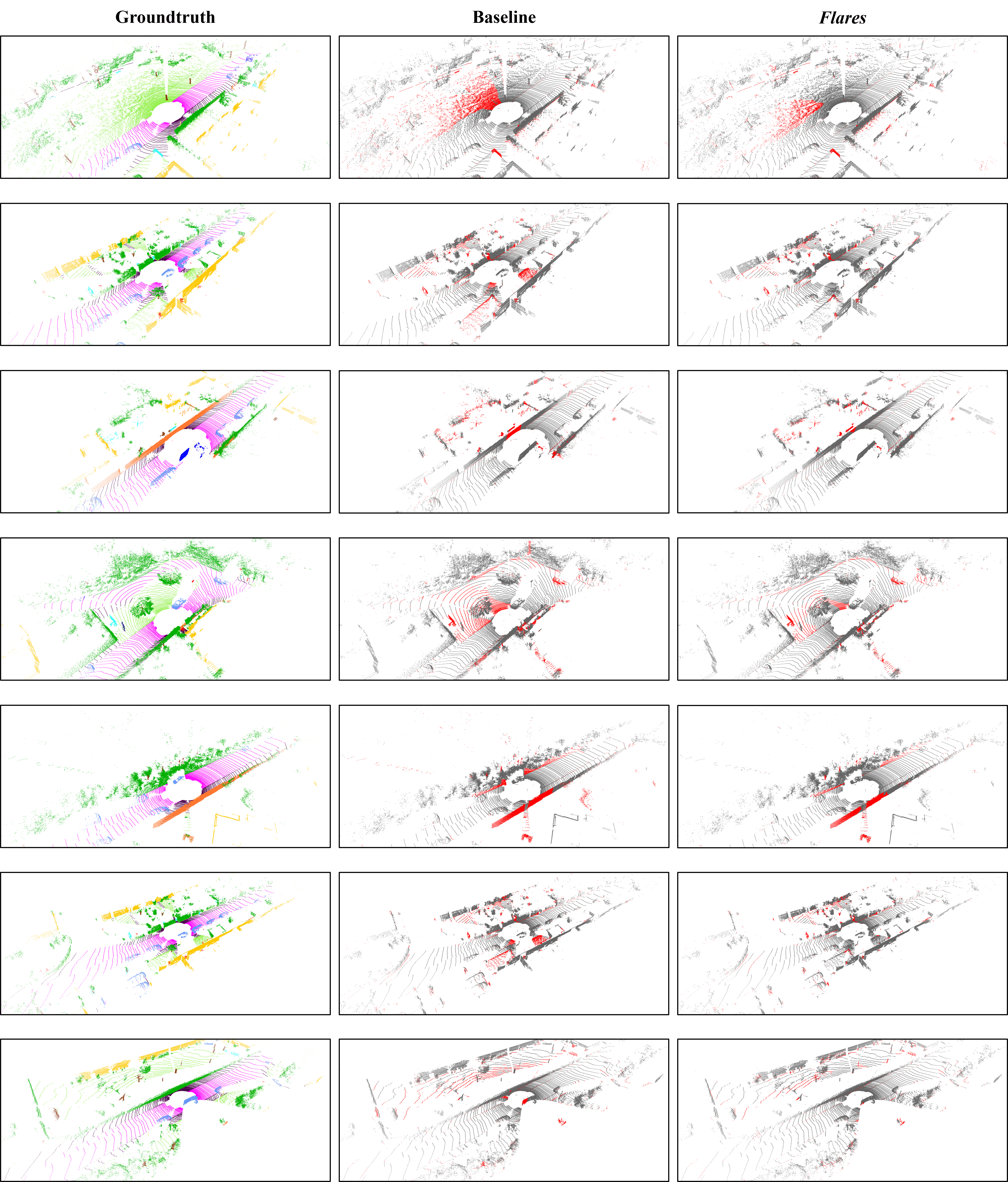}
    \vspace{1mm}
    \caption{Additional qualitative results of \textbf{CENet}~\cite{cheng2022cenet}: points in \textcolor{incorrect}{red} and \textcolor{correct}{gray} represent incorrect and correct predictions, respectively. The groundtruth image is color-coded to differentiate between classes.}
    \label{fig:supp_cenet_qualitative}
\end{figure*}
\section{Video Demo}
In addition to the figures, we have included two video demos per network tested (eight videos in total) in the supplementary materials to showcase a more comprehensive evaluation of our approach. Each video contains hundreds of frames, with each frame presenting three visualizations: groundtruth, baseline prediction, and \coolname{} prediction. Each visualization includes both top-down-view and range-view images of the LiDAR point cloud. For videos labeled with the suffix \textit{errormap}, predictions are displayed in a binary format, with incorrect predictions shown in \textcolor{incorrect}{red} and correct ones in \textcolor{correct}{gray}. %These video demos are accessible at the following link: \url{https://www.youtube.com/playlist?list=PLwXbFNuXzMAWhODt51hBwjjnj80y6wFi0}.
\clearpage
% 1. More qualitative results
% 3. Video demo

%% file: tables/supp_semantickitti.tex
\begin{table*}[h]
\vspace{-3mm}
\centering\scalebox{0.67}{
\begin{tabular}{r|c|cccccccccccccccccccc}
\toprule
\multicolumn{21}{c}{~\textbf{SemanticKITTI \textit{test} set}}
\\\midrule
 \textbf{Method~\small{(year)}}  & \rotatebox{0}{$\text{mIoU}$} & \rotatebox{0}{car} & \rotatebox{0}{bicy} & \rotatebox{0}{moto} & \rotatebox{0}{truc} & \rotatebox{0}{o.veh} & \rotatebox{0}{ped} & \rotatebox{0}{b.list} & \rotatebox{0}{m.list} & \rotatebox{0}{road} & \rotatebox{0}{park} & \rotatebox{0}{walk} & \rotatebox{0}{o.gro} & \rotatebox{0}{build} & \rotatebox{0}{fenc} & \rotatebox{0}{veg} & \rotatebox{0}{trun} & \rotatebox{0}{terr} & \rotatebox{0}{pole} & \rotatebox{0}{sign}
\\\midrule
RandLA-Net~\cite{hu2020randla}~\small{['20]} & $50.3$ & $94.0$ & $19.8$ & $21.4$ & $42.7$ & $38.7$ & $47.5$ & $48.8$ & $4.6$ & $90.4$ & $56.9$ & $67.9$ & $15.5$ & $81.1$ & $49.7$ & $78.3$ & $60.3$ & $59.0$ & $44.2$ & $38.1$\\
PolarNet~\cite{zhang2020polarnet}~\small{['20]} & $54.3$ & $93.8$ & $40.3$ & $30.1$ & $22.9$ & $28.5$ & $43.2$ & $40.2$ & $5.6$ & $90.8$ & $61.7$ & $74.4$ & $21.7$ & $90.0$ & $61.3$ & $84.0$ & $65.5$ & $67.8$ & $51.8$ & $57.5$\\
SqSegV3~\cite{xu2020squeezesegv3}~\small{['20]} & $55.9$ & $92.5$ & $38.7$ & $36.5$ & $29.6$ & $33.0$ & $45.6$ & $46.2$ & $\mathbf{20.1}$ & $91.7$ & $63.4$ & $74.8$ & $26.4$ & $89.0$ & $59.4$ & $82.0$ & $58.7$ & $65.4$ & $49.6$ & $58.9$\\
KPConv~\cite{thomas2019kpconv}~\small{['20]} & $58.8$ & \underline{$96.0$} & $32.0$ & $42.5$ & $33.4$ & $44.3$ & $61.5$ & $61.6$ & \underline{$11.8$} & $88.8$ & $61.3$ & $72.7$ & \underline{$31.6$} & $\mathbf{95.0}$ & $64.2$ & $84.8$ & $69.2$ & \underline{$69.1$} & $56.4$ & $47.4$\\
FusionNet~\cite{quan2021fusionnet}~\small{['20]} & $61.3$ & $95.3$ & $47.5$ & $37.7$ & $41.8$ & $34.5$ & $59.5$ & $56.8$ & $11.9$ & \underline{$91.8$} & $68.8$ & \underline{$77.1$} & $30.8$ & \underline{$92.5$} & $\mathbf{69.4}$ & $\mathbf{85.6}$ & \underline{$69.8$} & $68.5$ & \underline{$60.4$} & \underline{$66.5$}\\
% TornadoNet~\cite{gerdzhev2021tornado}~\small{['20]} & $63.1$ & $94.2$ & $55.7$ & $48.1$ & $40.0$ & $38.2$ & $63.6$ & $60.1$ & $34.9$ & $89.7$ & $66.3$ & $74.5$ & $28.7$ & $91.3$ & $65.6$ & $85.6$ & $67.0$ & $71.5$ & $58.0$ & $65.9$\\
AMVNet~\cite{liong2020amvnet}~\small{['20]} & $\mathbf{65.3}$ & $\mathbf{96.2}$ & $\mathbf{59.9}$ & \underline{$54.2$} & \underline{$48.8$} & \underline{$45.7$} & $\mathbf{71.0}$ & \underline{$65.7$} & $11.0$ & $90.1$ & $\mathbf{71.0}$ & $75.8$ & $\mathbf{32.4}$ & $92.4$ & \underline{$69.1$} & $\mathbf{85.6}$ & $\mathbf{71.7}$ & $\mathbf{69.6}$ & $\mathbf{62.7}$ & $\mathbf{67.2}$\\
\textcolor{orange}{\text{\faBolt}}\textbf{SalsaNext}\textcolor{orange}{\text{\faBolt}}~\cite{cortinhal2020salsanext}~\small{['20]}  &   \underline{$64.8$} &    $95.1$ &   \underline{$55.5$} &   $\mathbf{56.5}$ &     $\mathbf{60.1}$ &   $\mathbf{53.7}$ &   \underline{$69.6$} &   $\mathbf{74.1}$ &   $11.4$ &   $\mathbf{93.0}$ &   \underline{$68.9$} &   $\mathbf{78.9}$ &   $20.4$ &   $91.1$ &   $67.6$ &   $82.0$ &   $66.7$ &   $65.0$ &   $58.1$ &   $64.1$\\
\arrayrulecolor[rgb]{0.4,0.4,0.4} \hline
MPF~\cite{alnaggar2021mpf}~\small{['21]} & $55.5$ & $93.4$ & $30.2$ & $38.3$ & $26.1$ & $28.5$ & $48.1$ & $46.1$ & $18.1$ & $90.6$ & $62.3$ & $74.5$ & $30.6$ & $88.5$ & $59.7$ & $83.5$ & $59.7$ & $69.2$ & $49.7$ & $58.1$\\
KPRNet~\cite{kochanov2020kprnet}~\small{['21]} & $63.1$ & $95.5$ & $54.1$ & $47.9$ & $23.6$ & $42.6$ & $65.9$ & $65.0$ & $16.5$ & $\mathbf{93.2}$ & $\mathbf{73.9}$ & $\mathbf{80.6}$ & $30.2$ & $91.7$ & $68.4$ & $\mathbf{85.7}$ & \underline{$69.8$} & $71.2$ & $58.7$ & $64.1$\\
LiteHDSeg~\cite{razani2021lite}~\small{['21]} & $63.8$ & $92.3$ & $40.0$ & $55.4$ & $37.7$ & $39.6$ & $59.2$ & \underline{$71.6$} & $\mathbf{54.3}$ & $93.0$ & $68.2$ & $78.3$ & $29.3$ & $91.5$ & $65.0$ & $78.2$ & $65.8$ & $65.1$ & $59.5$ & \underline{$67.7$}\\
JS3C-Net~\cite{yan2021sparse}~\small{['21]} & $66.0$ & \underline{$95.8$} & \underline{$59.3$} & $52.9$ & \underline{$54.3$} & $46.0$ & $69.5$ & $65.4$ & $39.9$ & $88.9$ & $61.9$ & $72.1$ & $31.9$ & $\mathbf{92.5}$ & $\mathbf{70.8}$ & $84.5$ & \underline{$69.8$} & $67.9$ & \underline{$60.7$} & $\mathbf{68.7}$\\
Cylinder3D~\cite{zhu2021cylindrical}~\small{['21]} & $\mathbf{68.9}$ & $\mathbf{97.1}$ & $\mathbf{67.6}$ & $\mathbf{63.8}$ & $50.8$ & \underline{$58.5$} & $\mathbf{73.7}$ & $69.2$ & \underline{$48.0$} & $92.2$ & $65.0$ & $77.0$ & \underline{$32.3$} & $90.7$ & $66.5$ & \underline{$85.6$} & $\mathbf{72.5}$ & \underline{$69.8$} & $\mathbf{62.4}$ & $66.2$\\
\textcolor{orange}{\text{\faBolt}}\textbf{FIDNet}\textcolor{orange}{\text{\faBolt}}~\cite{zhao2021fidnet}~\small{['21]}  &  \underline{$67.4$} &    \underline{$95.8$} &   $56.7$ &   \underline{$60.7$} &     $\mathbf{58.1}$ &   $\mathbf{60.3}$ &   \underline{$72.5$} &   $\mathbf{72.9}$ &   $15.8$ &   $\mathbf{93.2}$ &   \underline{$69.2$} &   \underline{$79.9$} &   $\mathbf{34.2}$ &   \underline{$91.9$} &   \underline{$69.0$} &   $84.6$ &   $68.7$ &   $\mathbf{70.3}$ &   $59.9$ &   $66.9$\\
\arrayrulecolor[rgb]{0.4,0.4,0.4} \hline
Meta-RSeg~\cite{wang2022meta}~\small{['22]} & $61.0$ & $93.9$ & $50.1$ & $43.8$ & $43.9$ & $43.2$ & $63.7$ & $53.1$ & $18.7$ & $90.6$ & $64.3$ & $74.6$ & $29.2$ & $91.1$ & $64.7$ & $82.6$ & $65.5$ & $65.5$ & $56.3$ & $64.2$\\
PCSCNet~\cite{park2023pcscnet}~\small{['22]} & $62.7$ & $95.7$ & $48.8$ & $46.2$ & $36.4$ & $40.6$ & $55.5$ & $68.4$ & $\mathbf{55.9}$ & $89.1$ & $60.2$ & $72.4$ & $23.7$ & $89.3$ & $64.3$ & $84.2$ & $68.2$ & $68.1$ & $60.5$ & $63.9$\\
GFNet~\cite{qiu2022gfnet}~\small{['22]} & $65.4$ & \underline{$96.0$} & $53.2$ & $48.3$ & $31.7$ & $47.3$ & $62.8$ & $57.3$ & $44.7$ & $\mathbf{93.6}$ & $\mathbf{72.5}$ & $\mathbf{80.8}$ & \underline{$31.2$} & $\mathbf{94.0}$ & $\mathbf{73.9}$ & \underline{$85.2$} & $71.1$ & $69.3$ & $61.8$ & $68.0$\\
MaskRange~\cite{gu2022maskrange}~\small{['22]} & $66.1$ & $94.2$ & $56.0$ & $55.7$ & \underline{$59.2$} & $52.4$ & $67.6$ & $64.8$ & $31.8$ & $91.7$ & \underline{$70.7$} & $77.1$ & $29.5$ & $90.6$ & $65.2$ & $84.6$ & $68.5$ & $69.2$ & $60.2$ & $66.6$ \\
GASN~\cite{GASN}~\small{['22]} & $\mathbf{70.7}$ & $\mathbf{96.9}$ & $\mathbf{65.8}$ & \underline{$58.0$} & $\mathbf{59.3}$ & $\mathbf{61.0}$ & $\mathbf{80.4}$ & $\mathbf{82.7}$ & \underline{$46.3$} & $89.8$ & $66.2$ & $74.6$ & $30.1$ & \underline{$92.3$} & \underline{$69.6$} & $\mathbf{87.3}$ & $\mathbf{73.0}$ & $\mathbf{72.5}$ & $\mathbf{66.1}$ & $\mathbf{71.6}$\\
\textcolor{orange}{\text{\faBolt}}\textbf{CENet}\textcolor{orange}{\text{\faBolt}}~\cite{cheng2022cenet}~\small{['22]} &    \underline{$68.0$} &    $95.9$ &     \underline{$61.1$} &     $\mathbf{62.1}$ &   $57.2$ &     \underline{$59.0$} &     \underline{$77.2$} &     \underline{$74.2$} &     $12.2$ &  \underline{$92.2$} &     $69.9$ &     \underline{$78.7$} &   $\mathbf{32.9}$ &     $91.8$ &     $68.8$ &     $84.7$ &     \underline{$71.3$} &     \underline{$69.9$} &     \underline{$62.9$} &   \underline{$70.3$}
% \\
% \arrayrulecolor[rgb]{0.4,0.4,0.4} \hline
% \arrayrulecolor{gray}\hline

% \textcolor{orange}{\text{\faBolt}}\textbf{RangeViT}\textcolor{orange}{\text{\faBolt}}~\small{['23]}  &   $\mathbf{66.1}$ &   $\mathbf{95.6}$ &   $\mathbf{56.3}$ &   $\mathbf{60.5}$ &   $\mathbf{52.4}$ &   $\mathbf{57.1}$ &   $\mathbf{72.0}$ &   $\mathbf{69.7}$ &   $16.0$ &     $91.6$ &   $\mathbf{71.1}$ &   ${77.3}$ &     $\mathbf{32.7}$ &   $91.4$ &   $67.4$ &   $83.1$ &   $68.0$ &   $68.1$ &   $58.0$ &     $\mathbf{67.5}$
\\\midrule
\end{tabular}}
\caption{The class-wise IoU scores of \textbf{\textit{different}} LiDAR semantic segmentation approaches on the \textit{test} set of SemanticKITTI~\cite{behley2019semantickitti}. All IoU score are given in percentage (\%). All approaches are categorized by the year of publication. In each block, \textbf{bold} and \underline{underline} indicate the \textbf{best} and \underline{second best} result in the column.} 
\label{tab:supp_semantickitti_full}
\end{table*}

%% file: tables/supp_nuscenes_testset.tex
\begin{table*}[h]
\vspace{-3mm}
\centering\scalebox{0.72}{
\begin{tabular}{r|c|cccccccccccccccc}
\toprule
\multicolumn{18}{c}{~\textbf{nuScenes \textit{test} set}}
\\\midrule
 \textbf{Method~\small{(year)}}  & \rotatebox{0}{$\text{mIoU}$} & \rotatebox{0}{barrier} & \rotatebox{0}{bicy} & \rotatebox{0}{bus} & \rotatebox{0}{car} & \rotatebox{0}{const} & \rotatebox{0}{moto} & \rotatebox{0}{ped} & \rotatebox{0}{traffic.c} & \rotatebox{0}{trailer} & \rotatebox{0}{truck} & \rotatebox{0}{driv} & \rotatebox{0}{o.flat} & \rotatebox{0}{side} & \rotatebox{0}{terrain} & \rotatebox{0}{manm} & \rotatebox{0}{veg}
\\\midrule
PolarNet~\cite{zhang2020polarnet}~\small{['20]} & $ 69.4 $ & $  72.2 $ & $ 16.8 $ & $ 77.0 $ & $ 86.5 $ & $ 51.1 $ & $ 69.7 $ & $  64.8 $ & $ 54.1 $ & $ 69.7 $ & $ 63.5 $ & $ 96.6 $ & $ 67.1 $ & $ 77.7 $ & $ 72.1 $ & $ 87.1 $ & $ 84.5$\\
JS3C-Net~\cite{yan2021sparse}~\small{['21]} & $ 73.6  $ & $ 80.1  $ & $ 26.2 $ & \underline{$ 87.8 $} & $ 84.5 $ & $ 55.2 $ & $ 72.6 $ & $ 71.3 $ & $ 66.3 $ & $ 76.8 $ & \underline{$ 71.2 $} & $ 96.8 $ & $ 64.5 $ & $ 76.9 $ & $ 74.1 $ & $ 87.5 $ & $ 86.1$\\
AMVNet~\cite{liong2020amvnet}~\small{['20]}  & $77.3 $ & $  80.6 $ & \underline{$ 32.0 $} & $ 81.7 $ & $ 88.9 $ & \underline{$ 67.1 $} & \underline{$ 84.3 $} & $ 76.1 $ & \underline{$ 73.5 $} & $\mathbf{84.9}$ & $ 67.3 $ & \underline{$ 97.5 $} & $ 67.4 $ & $ 79.4 $ & \underline{$ 75.5 $} & $ \mathbf{91.5} $ & $ 88.7$\\
Cylinder3D~\cite{zhu2021cylindrical}~\small{['21]}  & $ 77.2 $ & \underline{$82.8$} & $ 29.8 $ & $ 84.3 $ & $ 89.4 $ & $ 63.0 $ & $ 79.3 $ & $ \mathbf{77.2} $ & $ 73.4 $ & $ 84.6 $ & $ 69.1 $ & $\mathbf{97.7}$ & \underline{$ 70.2 $} & $\mathbf{80.3} $ & \underline{$ 75.5 $} & $ 90.4 $ & $ 87.6$  \\
RangeFormer~\cite{kong2023rethinking}~\small{['23]} & $\mathbf{80.1}$ & $\mathbf{85.6}$ & $\mathbf{47.4}$ & $\mathbf{91.2}$ & $\mathbf{90.9}$ & $\mathbf{70.7}$ & $\mathbf{84.7}$ & \underline{$77.1$} & $\mathbf{74.1}$ & $83.2$ & $\mathbf{72.6}$ & \underline{$97.5$} & $\mathbf{70.7}$ & $79.2$ & $75.4$ & \underline{$91.3$} & \underline{$88.9$}\\
\textcolor{orange}{\text{\faBolt}}\textbf{CENet}\textcolor{orange}{\text{\faBolt}}~\small{['22]}   &  \underline{$77.5$} &    $82.2$ &   $31.1$ &   $85.2$ &     \underline{$90.7$} &   $65.1$ &   $82.5$ &   $74.9$ &   $72.3$ &   \underline{$84.7$} &   $70.5$ &   $97.4$ &   $69.5$ &   \underline{$79.6$} &   $\mathbf{76.1}$ &   $90.5$ &   $\mathbf{90.1}$
\\\bottomrule
\end{tabular}}
\caption{The class-wise IoU scores compared with \textbf{\textit{different}} LiDAR semantic segmentation approaches on the \textit{test} set of nuScenes~\cite{caesar2020nuscenes}. All IoU score are given in percentage (\%). Due to limited number of submission times, we only test \coolname{} with CENet~\cite{cheng2022cenet}. \textbf{Bold} and \underline{underline} indicate the \textbf{best} and \underline{second best} result in the column.} 
\label{table:supp_nuscenes_testset}
\end{table*}

%% file: tables/supp_semantickitti_notta.tex
\begin{table*}[h]
\vspace{-3mm}
\centering\scalebox{0.63}{
\begin{tabular}{l|c|c|cccccccccccccccccccc}
\toprule
\multicolumn{21}{c}{~\textbf{SemanticKITTI \textit{test} set}}
\\\midrule
 \textbf{Method~\small{(year)}} & \#params & \rotatebox{0}{$\text{mIoU}$} & \rotatebox{0}{car} & \rotatebox{0}{bicy} & \rotatebox{0}{moto} & \rotatebox{0}{truc} & \rotatebox{0}{o.veh} & \rotatebox{0}{ped} & \rotatebox{0}{b.list} & \rotatebox{0}{m.list} & \rotatebox{0}{road} & \rotatebox{0}{park} & \rotatebox{0}{walk} & \rotatebox{0}{o.gro} & \rotatebox{0}{build} & \rotatebox{0}{fenc} & \rotatebox{0}{veg} & \rotatebox{0}{trun} & \rotatebox{0}{terr} & \rotatebox{0}{pole} & \rotatebox{0}{sign}
\\\midrule
% \arrayrulecolor[rgb]{0.4,0.4,0.4} \hline
% \arrayrulecolor{gray}\hline
RangeFormer~\cite{kong2023rethinking}~\small{['23]} & 24.3M&   $\mathbf{69.5}$ &   $94.7$ &   \underline{$60.0$} &   $\mathbf{69.7}$ &   $\mathbf{57.9}$ &   $\mathbf{64.1}$ &   $72.3$ &   $\mathbf{72.5}$ &   $\mathbf{54.9}$ &     $90.3$ &   \underline{$69.9$} &   $74.9$ &     $\mathbf{38.9}$ &   $90.2$ &   $66.1$ &   $\mathbf{84.1}$ &   $68.1$ &   $\mathbf{70.0}$ &   $58.9$ &     $63.1$\\
LENet~\cite{ding2023lenet}~\small{['23]} & 4.7M&   $64.5$ &   $93.9$ &   $57.0$ &   $51.3$ &   $44.3$ &   $44.4$ &   $66.6$ &   $64.9$ &   \underline{$36.0$} &     $91.8$ &   $68.3$ &   $76.9$ &     $30.5$ &   $91.2$ &   $66.0$ &   $83.7$ &   $68.3$ &   $67.8$ &   $58.6$ &   $63.2$\\ 
TFNet~\cite{li2024tfnet}~\small{['24]} & -&   $66.1$ &   $94.3$ &   $\mathbf{60.7}$ &   $58.5$ &   $38.4$ &   $48.4$ &   \underline{$74.3$} &   $72.2$ &   $35.5$ &     $90.6$ &   $68.5$ &   $75.3$ &     $29.0$ &   $\mathbf{91.6}$ &   $67.3$ &   \underline{$83.8$} &   $\mathbf{71.1}$ &   $67.0$ &   $\mathbf{60.8}$ &   $\mathbf{68.7}$\\
\textcolor{orange}{\text{\faBolt}}\textbf{SalsaNext}\textcolor{orange}{\text{\faBolt}}~\small{['20]}  & 6.7M &   {$63.3$} &    {$94.7$} &   {$52.9$} &   {$55.7$} &     {$57.3$} &   {$50.2$} &   {$65.5$} &   {$70.9$} &   {$13.0$} &   $\mathbf{92.6}$ &   $69.0$ &   \underline{$77.7$} &   {$20.5$} &   {$90.4$} &   {$65.8$} &   $80.8$ &   {$65.0$} &   $63.4$ &   {$55.4$} &   {$62.4$} \\
\textcolor{orange}{\text{\faBolt}}\textbf{FIDNet}\textcolor{orange}{\text{\faBolt}}~\small{['21]}  & 6.1M &  {$65.1$} &    {$95.3$} &   $51.0$ &   {$57.0$} &     {$54.8$} &   \underline{$58.1$} &   {$68.1$} &   {$68.9$} &   $14.4$ &   \underline{$92.3$} &   {$68.3$} &   $\mathbf{78.0}$ &   {$32.3$} &   $\mathbf{91.6}$ &   $\mathbf{67.6}$ &   $83.7$ &   {$66.6$} &   \underline{$68.8$} &   {$55.1$} &   {$64.8$}\\
\textcolor{orange}{\text{\faBolt}}\textbf{CENet}\textcolor{orange}{\text{\faBolt}}~\small{['22]}  & 6.8M  &   \underline{$66.6$} &     $\mathbf{95.6}$ &     $58.5$ &     \underline{$61.6$} &   {$51.7$} &     {$50.2$} &     $\mathbf{74.5}$ &     \underline{$72.4$} &     $23.2$ &   {$91.4$} &     {$69.6$} &     {$77.1$} &   {$31.7$} &     {$91.1$} &     {$66.6$} &     \underline{$83.8$} &     \underline{$69.9$} &     $68.3$ &     \underline{$60.3$} &   $\mathbf{68.7}$\\
\textcolor{orange}{\text{\faBolt}}\textbf{RangeViT}\textcolor{orange}{\text{\faBolt}}~\small{['23]}    & 23.6M&   $66.1$ &   $\mathbf{95.6}$ &   $56.3$ &   $60.5$ &   $52.4$ &   $\mathbf{57.1}$ &   $72.0$ &   $69.7$ &   $16.0$ &     $91.6$ &   $\mathbf{71.1}$ &   ${77.3}$ &     \underline{$32.7$} &   $91.4$ &   \underline{$67.4$} &   $83.1$ &   $68.0$ &   $68.1$ &   $58.0$ &     $67.5$
\\\midrule
\end{tabular}}
\caption{The class-wise IoU scores compared with \textbf{\textit{latest range-view}} LiDAR semantic segmentation approaches on the \textit{test} set of SemanticKITTI~\cite{behley2019semantickitti}. All IoU score are given in percentage (\%). All approaches are evaluated \textbf{\textit{without test-time augmentation}}. \textbf{Bold} and {underline} indicate the \textbf{best} and {second best} result in the column.} 
\label{tab:supp_semantickitti_notta}
\end{table*}